\documentclass{article}

\usepackage[final, nonatbib]{neurips_2023}

\usepackage[utf8]{inputenc} %
\usepackage[T1]{fontenc}    %
\usepackage{xcolor}
\usepackage[square,numbers]{natbib}
\usepackage{textcomp}
\usepackage{float}
\usepackage{gensymb}
\usepackage[colorlinks=true, linkcolor=red, urlcolor=blue, citecolor=blue, anchorcolor=blue]{hyperref}
\usepackage{booktabs}       %
\usepackage{amsfonts}       %
\usepackage{nicefrac}       %
\usepackage{microtype}      %
\usepackage{mathtools}
\usepackage{lipsum}
\usepackage{graphicx}
\usepackage{cleveref}
\usepackage[shortlabels]{enumitem}
\usepackage{tikz}
\usepackage{wrapfig}
\usepackage{subcaption}
\usepackage{tabularx}

\usepackage{pdflscape}
\usepackage{bm}
\usepackage{titlesec}
\usepackage{amsmath, amsthm}
\usepackage{thmtools, thm-restate}
\usepackage{colortbl}
\usepackage{algorithm}
\usepackage{multirow}
\usepackage{amsthm,amssymb,bm,color}
\usepackage[etex=true,export]{adjustbox}

\DeclareMathOperator*{\argmax}{arg\,max}
\DeclareMathOperator*{\argmin}{arg\,min}

\usepackage[noend]{algpseudocode}
\usepackage{footnote}
\newcommand{\approxtext}[1]{\ensuremath{\stackrel{\text{#1}}{\approx}}}
\newcommand\data[1]{{\normalfont \texttt{#1}}}

\definecolor{green}{RGB}{34,139,34}

\title{\!FlatMatch: Bridging Labeled Data and Unlabeled Data with Cross-Sharpness for Semi-Supervised Learning}

\author{
	Zhuo Huang$^{1}$,  
	Li Shen$^{2}$,   
	Jun Yu$^3$,
	\textbf{Bo Han$^4$,
		Tongliang Liu$^{1, }\thanks{Corresponding to Tongliang Liu.}$} \\[1ex]
	\small{$^1$Sydney AI Centre, The University of Sydney;}
	\small{$^2$JD Explore Academy;}\\
	\small{$^3$Department of Automation, University of Science and Technology of China;}\\
	\small{$^4$Department of Computer Science, Hong Kong Baptist University}\\
	\texttt{zhua0420@uni.sydney.edu.au, mathshenli@gmail.com, harryjun@ustc.edu.cn,}\\
	\texttt{bhanml@comp.hkbu.edu.hk, tongliang.liu@sydney.edu.au}\\
}

\begin{document}

\maketitle

\begin{abstract}
	Semi-Supervised Learning (SSL) has been an effective way to leverage abundant unlabeled data with extremely scarce labeled data. However, most SSL methods are commonly based on instance-wise consistency between different data transformations. Therefore, the label guidance on labeled data is hard to be propagated to unlabeled data. Consequently, the learning process on labeled data is much faster than on unlabeled data which is likely to fall into a local minima that does not favor unlabeled data, leading to sub-optimal generalization performance. In this paper, we propose \textit{FlatMatch} which minimizes a cross-sharpness measure to ensure consistent learning performance between the two datasets. Specifically, we increase the empirical risk on labeled data to obtain a worst-case model which is a failure case that needs to be enhanced. Then, by leveraging the richness of unlabeled data, we penalize the prediction difference (\textit{i.e.}, cross-sharpness) between the worst-case model and the original model so that the learning direction is beneficial to generalization on unlabeled data. Therefore, we can calibrate the learning process without being limited to insufficient label information. As a result, the mismatched learning performance can be mitigated, further enabling the effective exploitation of unlabeled data and improving SSL performance. Through comprehensive validation, we show FlatMatch achieves state-of-the-art results in many SSL settings. Our code is available at \href[]{https://github.com/tmllab/2023_NeurIPS_FlatMatch}{https://github.com/tmllab/2023\_NeurIPS\_FlatMatch}.
\end{abstract}



\section{Introduction}
\label{sec:introduction}
Training deep models~\cite{he2016deep, lecun2015deep} not only requires countless data but also hungers for label supervision which commonly consumes huge human laboring and unaffordable monetary costs. To ease the dependency on labeled data, semi-supervised learning (SSL)~\cite{bennett1999semi, chapelle2009semi, grandvalet2005semi} has been one of the most effective strategies to exploit abundant unlabeled data by leveraging scarce labeled data. Traditional SSL commonly leverages unlabeled data by analyzing their manifold information. For example, semi-supervised support vector machine~\cite{bennett1999semi, chapelle2008optimization} finds a classifier that crosses the low-density region based on large-margin theorem and label-propagation~\cite{wang2007label, zhou2004learning} computes an affinity matrix in the input space to propagate the labels to their neighbor unlabeled data. Due to the computational burden of exploring manifold knowledge, recent advances in SSL benefit from sophisticated data augmentation techniques~\cite{bai2022rsa, cubuk2019autoaugment, cubuk2020randaugment, xie2020unsupervised, zhang2017mixup}, they usually enforce predictive consistency between the original inputs and their augmented variants~\cite{xie2020self, berthelot2019mixmatch, berthelot2019remixmatch, sohn2020fixmatch, wang2023freematch, wu2023conditional, xu2021dash, zhang2021flexmatch}, meanwhile using pseudo labels~\cite{lee2013pseudo, pham2020meta, rizve2021in, bai2021me} to guide the learning on unlabeled data.

\begin{figure*}[t]
	\begin{minipage}[t]{0.32\textwidth}
		\centering
		\text{\small \ 2D contours of labeled data}
		\includegraphics[width=\linewidth]{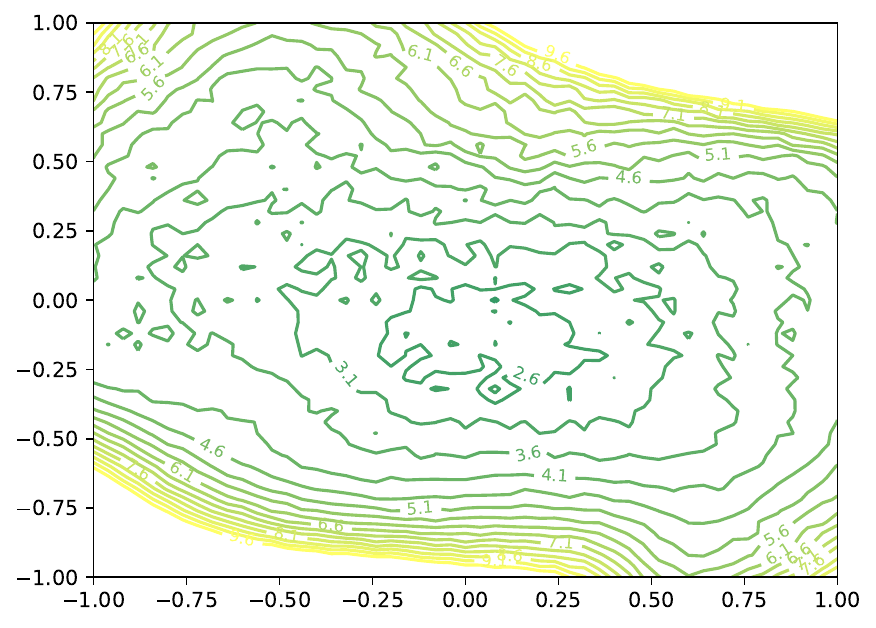}
	\end{minipage}
	\begin{minipage}[t]{0.32\textwidth}
		\centering
		\text{\small \ 2D contours of unlabeled data}
		\includegraphics[width=\linewidth]{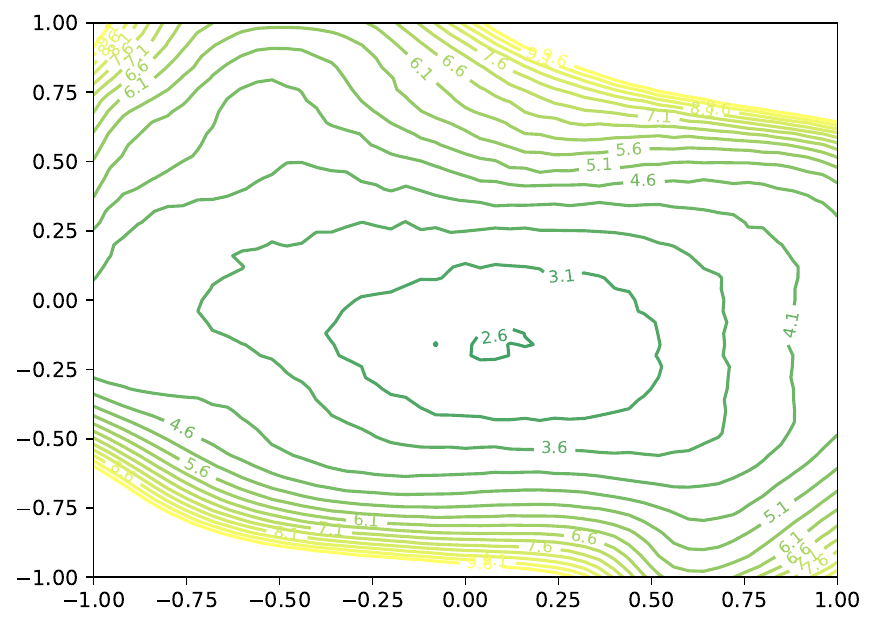}
	\end{minipage}
	\begin{minipage}[t]{0.35\textwidth}
		\centering
		\text{\small1D loss curves}
		\raisebox{0.01\height}{\includegraphics[width=\linewidth]{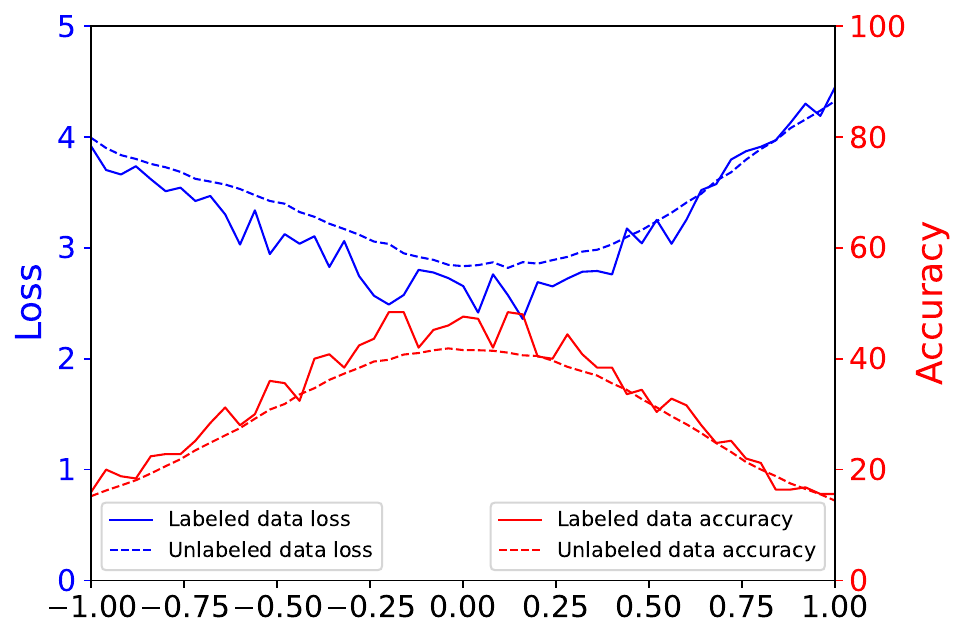}}
	\end{minipage}
	\begin{minipage}[t]{0.32\textwidth}
		\centering
		\includegraphics[width=\linewidth]{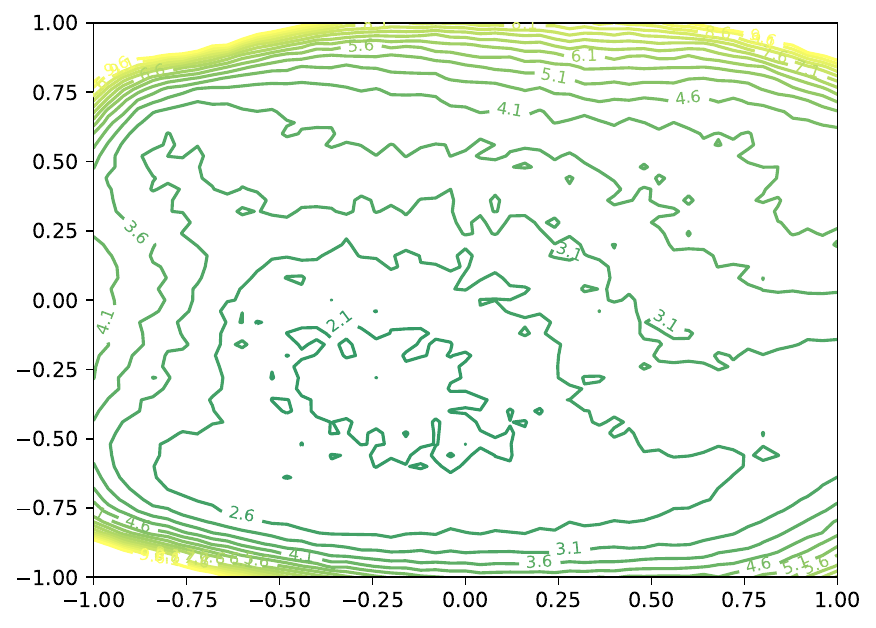}
	\end{minipage}
	\begin{minipage}[t]{0.32\textwidth}
		\centering
		\includegraphics[width=\linewidth]{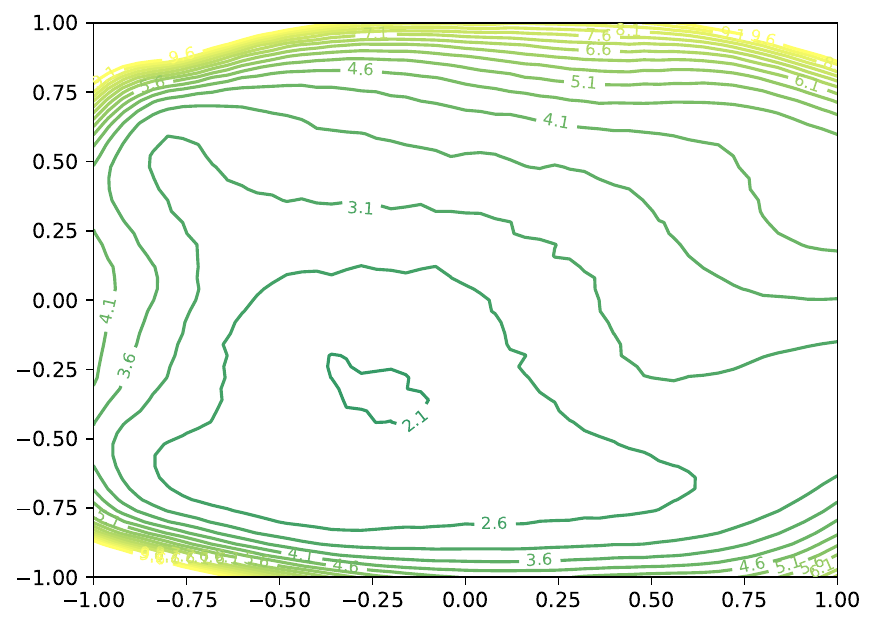}
	\end{minipage}
	\begin{minipage}[t]{0.35\textwidth}
		\centering
		\includegraphics[width=\linewidth]{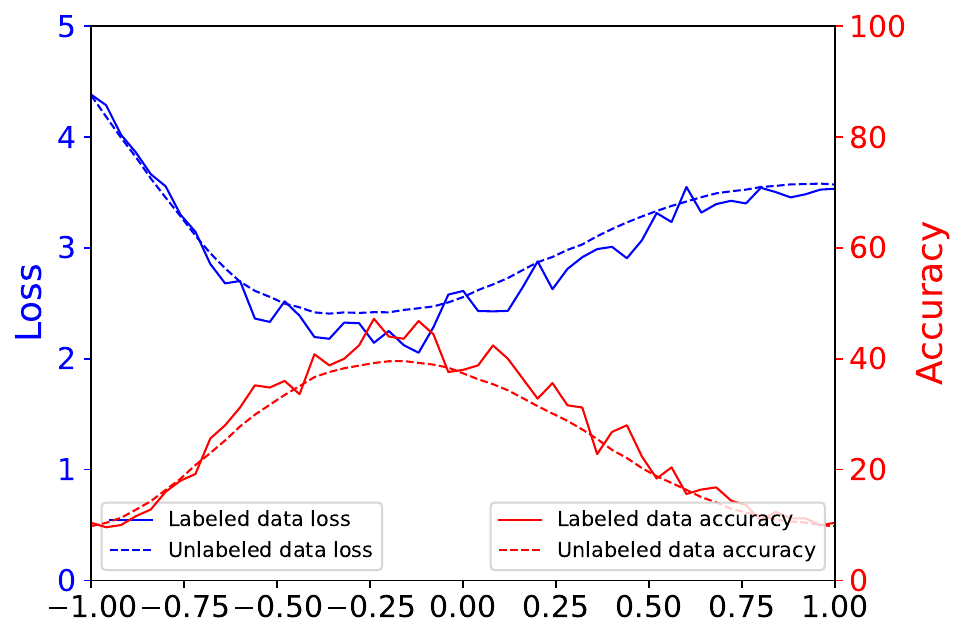}
	\end{minipage}
	\caption{\small Loss landscapes of labeled data and unlabeled data obtained simultaneously from training using FixMatch~\cite{sohn2020fixmatch} on CIFAR10 with 250 labels per class. The first row and second row show the results obtained from epoch 60 and epoch 150, respectively. The first column and second column show the 2D loss contours of labeled data and unlabeled data, respectively, and the last column shows the 1D loss curves.}
	\label{fig:motivation}
\end{figure*}

However, as most of the existing methods only apply instance-wise consistency on each example with its own transformation, labeled data and unlabeled data are  disconnected during training. Hence, the label guidance cannot be sufficiently propagated from the labeled set to the unlabeled set, causing the learning process on labeled data is much faster than on unlabeled data. As a result, the learning model could be misled to a local minima that cannot favor unlabeled data, leading to a non-negligible generalization mismatch. As generalization performance is closely related to the flatness of loss landscape~\cite{chaudhari2019entropy, foret2020sharpness, keskar2017large, neyshabur2017exploring}, we plot the loss landscapes of labeled data and unlabeled data using FixMatch~\cite{sohn2020fixmatch} in Fig.~\ref{fig:motivation}\footnote{Note that the loss curves cannot be directly generated using the raw data from training, because networks can fit all examples perfectly without showing any generalization evidence. Therefore, we load the data for plotting using slight data augmentation, \textit{i.e.}, Random Rotation with $90^{\circ}$, which is very commonly used in SSL.}. We have two major findings: 1) The loss landscape of labeled data is extremely sharp, but the loss curve of unlabeled data is quite flat. 2) Moreover, the optimal loss value of labeled data is much lower than that of unlabeled data. Such two intriguing discoveries reveal the \textbf{essence} of SSL: \textit{The learning on scarce labeled data convergences faster with lower errors than on unlabeled data, but it is vulnerable to perturbations and has an unstable loss curvature which rapidly increases as parameters slightly vary. Therefore, the abundant unlabeled data are leveraged so that SSL models are fitted to a wider space, thus producing a flatter loss landscape and generalizing better than labeled data}. Based on the analysis, existing methods that use instance-wise consistency with mismatched generalization performance could have two non-trivial pitfalls: 
\begin{itemize}
	\vspace{-2mm}
	\setlength\itemsep{0em}
	\item[--] The label supervision might not be sufficiently leveraged to guide the learning process of unlabeled data.
	\item[--] The richness brought by unlabeled data remains to be fully exploited to enhance the generalization performance of labeled data.
\end{itemize}

As seeking the connection in input space is quite prohibitive, we resort to exploring the parameter space and propose \textit{FlatMatch} which targets to encourage consistent learning performance between labeled data and unlabeled data, such that SSL models can obtain strong generalization ability without being limited by insufficient label information. Specifically, as the loss landscape on labeled data is quite unstable, we aim to mitigate this problem through applying an adversarial perturbation~\cite{foret2020sharpness, kwon2021asam, wu2020adversarial, zhao2022penalizing, zhao2022ss} to the model parameters so that the worst failure case can be found. Further, we enforce the worst-case model and the original model to achieve an agreement on unlabeled data via computing their prediction differences which are dubbed \textit{cross-sharpness}. By minimizing such a cross-sharpness measure, the richness of unlabeled data can be fully utilized to calibrate the learning direction. In turn, the label information can be activated for producing accurate pseudo labels, thus successfully bridging the two datasets to enable improved SSL performance.

Furthermore, the proposed FlatMatch is easy to implement and can be adapted to complement many popular SSL methods~\cite{cao2022openworld, chen2022debiased, chen2023softmatch, guo2022class}. Although computing the proposed cross-sharpness normally requires an additional back-propagation step, we propose an efficient strategy so that our FlatMatch can be processed without extra computational burden. Through extensive experiments and analyses on various benchmark datasets, we show that our FlatMatch achieves huge performance improvement compared to some of the most powerful approaches. Specifically, on CIFAR100 dataset with 2500 labeled examples, our FlatMatch surpasses the foremost competitor so far with 1.11\% improvement. In general, our contribution can be summarized into three points:
\begin{itemize}
	\vspace{-2mm}
	\setlength\itemsep{0em}
	\item[--] We identify a generalization mismatch of SSL due to the disconnection between labeled data and unlabeled data, which leads to two critical flaws that remain to be solved (Section~\ref{sec:introduction}).
	\item[--] We propose FlatMatch which addresses these problems by penalizing a novel cross-sharpness that helps improve the generalization performance of SSL (Section~\ref{sec:corss_sharpness}).
	\item[--] We reduce the computational cost of FlatMatch by designing an efficient implementation (Section~\ref{sec:efficient}).
	\item[--] Extensive experiments are conducted to fully validate the performance of FlatMatch, which achieves state-of-the-art results in several scenarios (Section~\ref{sec:experiments}).
\end{itemize}




\section{Related Works}
\label{sec:related_works}
SSL has been developed rapidly since the booming trend of deep learning~\cite{he2016deep, lecun2015deep, goodfellow2014generative}. We roughly introduce it through three stages, namely traditional SSL, data-augmented SSL, and open-set SSL.

\textbf{Traditional SSL:} One of the first SSL methods is Pseudo Label~\cite{lee2013pseudo} which has been very effective for training neural networks in a semi-supervised manner, which relies on the gradually improved learning performance to select confident samples~\cite{xia2021sample,xia2023combating}. Another classical method is Co-training~\cite{blum1998combining} which utilizes two separate networks to enforce consistent predictions on unlabeled data so that the classification can be more accurate. Inspired by co-training, many SSL approaches are motivated to conduct consistency regularization. Temporal ensembling~\cite{laine2016temporal} proposes to leverage historical model predictions to ensemble an accurate learning target that can be used to guide the learning of the current model. Mean Teacher~\cite{tarvainen2017mean} further suggests such an ensemble can be operated on the model weights to produce a teacher model that can provide improved learning targets. Instead of improving the teacher model, Noisy student~\cite{xie2020self} finds that adding noise to the student model can also improve SSL.

\textbf{Data-Augmented SSL:} Thanks to the development of data augmentation techniques and self-supervised learning~\cite{he2020momentum, hong2022semantic, wang2022exploring, wang2022mosaic}, further improvements in SSL are achieved by enforcing consistency between the original data and their augmented variants. VAT~\cite{miyato2018virtual} proposes to add adversarial perturbations~\cite{goodfellow2014generative, goodfellow2015explaining} to unlabeled data. By using the pseudo-label guidance to leverage such adversarial unlabeled data, the model can be more robust to strong input noises, thus showing better performance on the test set. MixMatch~\cite{berthelot2019mixmatch} employs the MixUp~\cite{zhang2017mixup} technique to interpolate between the randomly shuffled examples which largely smooths the input space as well as the label space, further enhancing the generalization result. Moreover, FixMatch~\cite{sohn2020fixmatch} is based on strong augmentation such as AutoAugment~\cite{cubuk2019autoaugment} and RandAugment~\cite{cubuk2020randaugment} to utilize the predictions of weakly augmented data to guide the learning of strongly augmented data, which can achieve near supervised learning performance on CIFAR10. Recently, Dash~\cite{xu2021dash}, FlexMatch~\cite{zhang2021flexmatch}, and FreeMatch~\cite{wang2023freematch} advanced SSL through investigating the threshold strategy for pseudo labels. They analyze the training dynamic of SSL and design various approaches to enhance the selection of unlabeled data, which effectively boosts the SSL performance to a new level.

\textbf{Open-Set SSL:} Moreover, another branch of SSL studies the realistic scenario where unlabeled data contains out-of-distribution (OOD) data~\cite{huang2021universal, huang2022they, he2022not, he2022towards, wang2022watermarking, wang2023doe, wang2023learning}. The common goal is to alleviate the influence of OOD such that the learning process of SSL will not be misled. Several studies propose many techniques that can achieve this goal, such as distillation~\cite{chen2020semi}, meta-learning~\cite{guo2020safe}, large-margin regularization~\cite{cao2022openworld}, content-style disentanglement~\cite{huang2023harnessing}, and consistency regularization can also help~\cite{saito2021openmatch}. But in this paper, we mainly focus on the common assumption that labeled data and unlabeled are sampled independently and identically.

\section{Preliminary: Improving Generalization via Penalizing Sharpness}
\label{sec:sam}
Sharpness-Aware Minimization (SAM)~\cite{foret2020sharpness, kwon2021asam, liu2022towards, wu2020adversarial, zhang2022ga} which focuses on optimizing the sharp points from parameter space so that the training model can produce a flat loss landscape. Specifically, given a set of fully-labeled data $\mathcal{D}^l=\{(x^l_i, y^l_i)\}_{i=1}^n$ containing $n$ data points $(x^l_i, y^l_i)$, SAM seeks to optimize a model parameterized by $\theta\in\Theta$ to fit the training dataset $\mathcal{D}^l$ so that $\theta$ can generalize well on a test set $\mathcal{D}^{te}$ drawn independently and identically as $\mathcal{D}^l$. Such an optimization process is normally conducted via minimizing the empirical risk $\mathcal{L}_{l}(\theta)=\frac{1}{n}\sum_{i=0}^n\ell_{ce}(\theta; g_{\theta}(x_i^l), y^l_i)$ which is realized by the cross-entropy loss $\ell_{ce}$ computed between the label prediction $f^l=g_{\theta}(x_i^l)$ and the class label $y^l_i$. To find the sharp points that maximally increase the empirical risk, SAM applies a weight perturbation $\epsilon\in\Theta$ to the model parameters $\theta$ and conducts the following inner maximization:
\begin{equation}
	\epsilon^*(\theta):=\argmax_{\|\epsilon\|_p\le\rho} \mathcal{L}_{l}(\theta+\epsilon)\approx\argmax_{\|\epsilon\|_p\le\rho}\epsilon^{\top}\nabla_{\theta}\mathcal{L}_{l}(\theta)\approxtext{$p=2$}\rho\frac{\nabla_{\theta}\mathcal{L}_{l}(\theta)}{\|\nabla_{\theta}\mathcal{L}_{l}(\theta)\|_2},
	\label{eq:sam_max}
\end{equation} 
where $\epsilon^*$ denotes the optimal perturbation which can be estimated using the gradient information of mini-batch inputs, $\rho$ restricts the perturbation magnitude of $\epsilon$ within a $\ell_p$-ball, and the approximation is given by the dual norm problem~\cite{foret2020sharpness}.

By perturbing the original model $\theta$ with $\epsilon^*$, we can obtain the \textit{sharpness} measure from SAM: $sharpness:=\mathcal{L}_{l}(\theta+\epsilon^*) - \mathcal{L}_{l}(\theta)$ which indicates the how quickly loss changes under the worst-case model perturbation $\epsilon^*$. To combine the sharpness term with empirical risk minimization (ERM), the SAM objective can be differentiated as follows:
\begin{equation}
	\begin{aligned}
		\nabla_{\theta}\mathcal{L}_{l}^{SAM}:=&\nabla_{\theta}\left[\mathcal{L}_{l}(\theta+\epsilon^*(\theta)) - \mathcal{L}_{l}(\theta)\right] + \mathcal{L}_{l}(\theta)
		\approx\nabla_{\theta}\mathcal{L}_{l}(\theta+\epsilon^*(\theta))\\
		=&\frac{d(\theta+\epsilon^*(\theta))}{d\theta}\nabla_{\theta}\mathcal{L}_{l}(\theta)|_{\theta+\epsilon^*(\theta)}=\nabla_{\theta}\mathcal{L}_{l}(\theta)|_{\theta+\epsilon^*(\theta)}+o(\theta),
		\label{eq:sam_obj}
	\end{aligned}
\end{equation}
where the last second-order term $o(\theta)$ can be discarded without significantly influencing the approximation. For detailed derivation and analysis, we refer to the original paper and related studies~\cite{andriushchenko2022towards, foret2020sharpness, zhao2022penalizing}. Intuitively, the gradient of SAM is computed on the worst-case point $\theta+\epsilon^*$, then such a gradient is utilized to update the original parameter $\theta$. As a result, the model can produce a more flat loss landscape than ERM which would not change drastically, and being robust along the sharpest direction $\epsilon^*$ among its $\ell_p$-norm neighbor.

Since the obtained flatness property has been demonstrated to have many benefits for generalization, such as being resistant to distribution shift~\cite{cha2021swad, izmailov2018averaging, wang2023sharpness, huang2023robust}, robustness against adversarial attack~\cite{wu2020adversarial, stutz2021relating}, and effectiveness under label noise~\cite{foret2020sharpness, bai2021understanding, huang2023paddles, kang2023unleashing, wu2023making, yuan2023late}, SAM has inspired many research progresses. However, SAM is only conducted under a fully-supervised setting, and whether or how can it be leveraged to improve learning with unlabeled data is still undiscovered. Next, we carefully introduce the proposed FlatMatch which improves the generalization performance of SSL by bridging labeled data and unlabeled data with a novel cross-sharpness.

\section{FlatMatch: Semi-Supervised Learning with Cross-Sharpness}
In this section, we carefully introduce our FlatMatch. First, we describe the SSL setting in a generalized way. Then, we demonstrate a novel regularization from FlatMatch which is dubbed cross-sharpness. Finally, we explain its optimization and design an efficient implementation.

\label{sec:flatmatch}
\subsection{General Semi-Supervised Learning}
In SSL, we are commonly given a small set of labeled data $\mathcal{D}^l=\{(x^l_i, y^l_i)\}_{i=1}^n$ containing $n$ labeled examples $(x^l_i, y^l_i)$ and a large unlabeled data $\mathcal{D}^u=\{x^u_i\}_{i=1}^m$ containing $m$ ($m\gg n$) unlabeled examples $x^u_i$ which are drawn independently and identically. Similar to previous notations, we aim to optimize a deep model $\theta\in\Theta$ in a semi-supervised manner so that $\theta$ can perform well on i.i.d sampled test set $\mathcal{D}^{te}$. The general objective for SSL is as follows:
\begin{equation}
	\min_{\theta}\mathcal{L}_{ssl}=\min_{\theta}\mathcal{L}_{l}+\mathcal{L}_{u}=\min_{\theta}\frac{1}{n}\sum_{i=0}^n\ell_{ce}(\theta;g_{\theta}(x^l_i), y^l_i)+\frac{1}{m}\sum_{i=0}^m\mathbb{I}(\hat{y}_i>\tau)\ell_{d}(\theta;g_{\theta}(\mathcal{A}(x^u_i)), \hat{y}_i),
	\label{eq:semi_obj}
\end{equation}
where the first term is similar to Eq.~\ref{eq:sam_max} and denotes the empirical risk under the label supervision, the second term indicates the semi-supervised regularization for exploiting unlabeled data, the function $\ell_{d}(\cdot)$ denotes the loss criterion for unlabeled data, such as KL-divergence or cross-entropy loss, $\mathcal{A}(\cdot)$ stands for an augmentation function that transforms the original input $x_i$ to its augmented variants, $\hat{y}_i$ represents the unsupervised learning target which can be a pseudo label or a guiding signal from a teacher model, and an indexing function $\mathbb{I}(\cdot)$ is commonly used to select the confident unlabeled data if their learning target surpasses a predefined threshold $\tau$. Our FlatMatch replaces the second term with a cross-sharpness regularization as described below.

\begin{figure}
	\centering
	\includegraphics[width=0.6\linewidth]{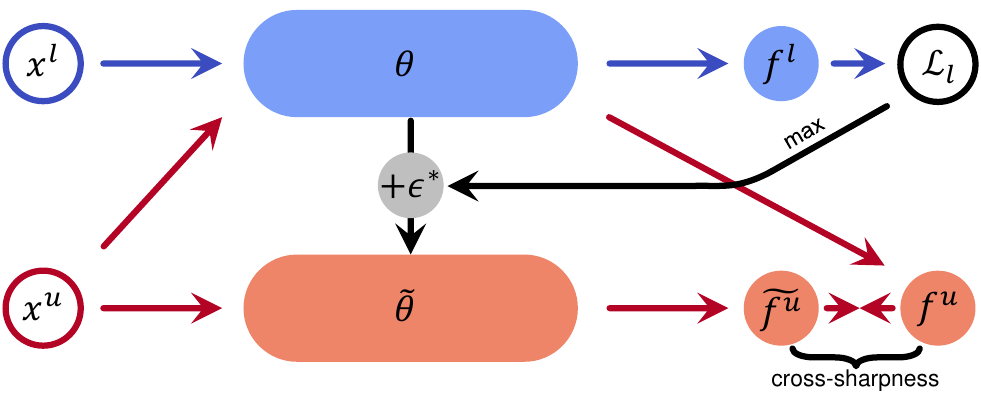}
	\caption{\small Illustration of FlatMatch. The blue arrows and red arrows denote the learning flows related to labeled data and unlabeled data, respectively; and the black arrows indicate the computation of the worst-case model.}
	\label{fig:flatmatch}
	\vspace*{-5mm}
\end{figure}

\subsection{The Cross-Sharpness Regularization}
\label{sec:corss_sharpness}
The computation of cross-sharpness can be briefly illustrated in Fig.~\ref{fig:flatmatch}. Particularly, to solve the sharp loss landscape problem, as demonstrated in Section~\ref{sec:introduction}, we penalize the cross-sharpness regularization on each unlabeled example using a worst-case model derived from maximizing the loss on labeled data. Formally, such optimization is shown as follows:
\begin{equation}
	\min_{\theta}\mathcal{R}_{\textit{X-sharp}}:=\min_{\theta}\frac{1}{m}\sum_{x^u\in\mathcal{D}^u}\ell_{d}(g_{\tilde{\theta}}(x^u), g_{\theta}(x^u)),\  \text{where} \ \tilde{\theta}=\theta+\argmax_{\|\epsilon\|_p\le\rho}\mathcal{L}_{l},
	\label{eq:cross_sharpness}
\end{equation}
in which $g_{\theta}(\cdot)$ indicates the forward propagation using the parameter $\theta$ whose prediction result is $f=g_{\theta}(\cdot)$, and $\tilde{\theta}$ stands for the worst-case model that maximally increases the empirical risk on labeled data. Here we drop the subscript $i$ from each datum $x$ for simplicity. By meaning ``cross'', the sharpness is not obtained by revisiting labeled data as SAM does, instead, we compute the sharpness by leveraging abundant unlabeled data to take full advantage of their generalization potential.


In detail, we leverage the original model $\theta$ to conduct a first propagation step as $f^l= g_{\theta}(x^l); f^u= g_{\theta}(x^u)$, where $f^l$ and $f^u$ are label predictions from labeled data and unlabeled data, respectively. Then, the empirical risk on labeled data is computed as $\mathcal{L}_{l}=\sum_{x^l\in\mathcal{D}^l}\ell_{ce}(\theta; g_{\theta}(x^l), y^l)$. Following Eq.~\ref{eq:sam_max}, we maximize the empirical risk $\mathcal{L}_{l}$ to obtain the weight perturbation as $\epsilon^*:=\rho\frac{\nabla_{\theta}\mathcal{L}_{l}(\theta)}{\|\nabla_{\theta}\mathcal{L}_{l}(\theta)\|_2}$, further having the worst-case model $\tilde{\theta}=\theta+\epsilon^*$. Moreover, we conduct a second propagation step by passing the unlabeled data to $\tilde{\theta}$ and have $\tilde{f^u}= g_{\tilde{\theta}}(x^u)$. The second unlabeled data prediction $\tilde{f^u}$ is combined with the first prediction $f^u$ obtained before to compute the cross-sharpness in Eq.~\ref{eq:cross_sharpness}\footnote{Note that there is a slight difference between the sharpness from SAM and our cross-sharpness: The former one is measured by the change of cross-entropy loss under model perturbation. But in SSL, the pseudo labels of unlabeled data is not accurate enough compared to labeled data, so we use the difference between label predictions of $\theta$ and $\tilde{\theta}$ to measure our cross-sharpness.}.


\textbf{Why does crossing the sharpness work?} As shown by many theoretical signs of progress, the generalization error of SSL is largely dependent on the number of labels~\cite{ben2008does, maximov2018rademacher}. Especially in recently proposed barely-supervised learning~\cite{sohn2020fixmatch}, the generalization error would be enormous. Although such a result may be pessimistic, unlabeled data can still be helpful. It is shown that when two distinct hypotheses on labeled data are co-regularized to achieve an agreement on unlabeled data, the generalization error can be reduced, as we quote: ``The reduction is proportional to the difference between the representations of the labeled data in the two different views, where the measure of difference is determined by the unlabeled data''~\cite{rosenberg2007rademacher}. Intuitively, disconnecting labeled data and unlabeled data during training may be sub-optimal. To achieve better theoretical performance, the SSL model should focus on \textit{learning something from unlabeled data that are \textbf{not contained} by labeled data}. In our FlatMatch, we can achieve this goal by enforcing $\theta$ and $\tilde{\theta}$, which produces the maximal loss difference on labeled data, to be consistent in classifying unlabeled data. Therefore, FlatMatch can be more effective than other instance-wise consistent methods by bridging the knowledge of labeled data and unlabeled data for training.

\renewcommand{\algorithmicrequire}{\textbf{Input:}}
\renewcommand{\algorithmicensure}{\textbf{Output:}}
\begin{algorithm}[t]
	\small
	\caption{\small FlatMatch and FlatMatch-e}
	\label{alg:flatmatch}
	\begin{algorithmic}[1]
		\Require Labeled dataset $\mathcal{D}^l$, unlabeled dataset $\mathcal{D}^u$, model $\theta$, memory buffer $\mathcal{M}$ for storing historical gradients, and EMA factor $\alpha$ for updating buffer.
		\For{$t \in 0,1, \ldots,T-1$}
		\If {Efficient training}
		\Comment{\textit{\color{black!60} FlatMatch-e}}
		\State Compute optimal perturbation as $\epsilon^*=\rho\frac{\mathcal{M}^{t}}{\|\mathcal{M}^{t}\|_2}$;
		\Else
		\Comment{\textit{\color{black!60} FlatMatch}}
		\State Compute gradient as $\nabla_{\theta}\mathcal{L}_{l}(\theta)=\nabla_{\theta}\sum_{x^l\in\mathcal{D}^l}\ell_{ce}(\theta; g_{\theta}(x^l), y^l)$; \Comment{\textit{\color{black!60} First propagation step}}
		\State  Compute optimal perturbation as $\epsilon^*:=\rho\frac{\nabla_{\theta}\mathcal{L}_{l}(\theta)}{\|\nabla_{\theta}\mathcal{L}_{l}(\theta)\|_2}$;
		\EndIf
		\State Obtain worst-case model as $\tilde{\theta}=\theta+\epsilon^*$;
		\State Compute cross-sharpness via Eq.~\ref{eq:cross_sharpness};
		\State Optimizing $\theta$ via Eq.~\ref{eq:optimization}, meanwhile save the gradient $\nabla_{\theta}\mathcal{L}_{l}(\theta)$;
		\Comment{\textit{\color{black!60} Second propagation step}}
		\State Update memory buffer as $\mathcal{M}^{t+1} = (1-\alpha)\times\mathcal{M}^{t}+\alpha\times\nabla_{\theta}\mathcal{L}_{l}(\theta)$;
		\EndFor
	\end{algorithmic}
\end{algorithm}

\subsection{Optimization and Efficient Implementation}
\label{sec:efficient}
To this end, we solve the optimization of FlatMatch by substituting $\mathcal{L}_{u}$ in Eq.~\ref{eq:semi_obj} with $\mathcal{R}_{\textit{X-sharp}}$ in Eq.~\ref{eq:cross_sharpness}:
\begin{equation}
	\theta:=\argmin_{\theta}\mathcal{L}_l(\theta)+\mathcal{R}_{\textit{X-sharp}}(\theta+\epsilon^*(\theta))\approx\nabla_{\theta}\mathcal{L}_{l} + \nabla_{\theta}\mathcal{R}_{\textit{X-sharp}}(\theta)|_{\theta+\rho\frac{\nabla_{\theta}\mathcal{L}_{l}(\theta)}{\|\nabla_{\theta}\mathcal{L}_{l}(\theta)\|_2}}.
	\label{eq:optimization}
\end{equation}
Particularly, the second gradient is obtained on the worst-case point $\theta+\rho\frac{\nabla_{\theta}\mathcal{L}_{l}(\theta)}{\|\nabla_{\theta}\mathcal{L}_{l}(\theta)\|_2}$, then it is applied on the original model $\theta$ for gradient descent. As we can see, such optimization requires an additional back-propagation on labeled data, which doubles the computational complexity. Fortunately, our FlatMatch can be implemented efficiently with no extra computational burden than the baseline stochastic gradient descent (SGD) optimizer, which is dubbed FlatMatch-e. The implementation is summarized in Algorithm~\ref{alg:flatmatch}.

Although many efficient methods for implementing SAM have already been proposed~\cite{du2022efficient, du2022sharpness, liu2022towards, zhao2022ss}, they commonly need extra regularization or gradient decomposition, which are not harmonious to SSL and would complicate our method. In our scenario, FlatMatch has a critical difference from SAM regarding the two propagation steps: The gradients from the first step and second step are not computed on the same batch of data, \textit{i.e.}, they are crossed from labeled data to unlabeled data. Since in SSL, each labeled batch and unlabeled batch are randomly coupled, we are allowed to use the gradient computed from the last batch of labeled data to obtain the cross-sharpness from the current batch of unlabeled data. Even better, we can use exponential moving average (EMA)~\cite{he2020momentum, laine2016temporal, tarvainen2017mean} to stabilize the gradient so that our cross-sharpness can be computed accurately. Next, we conduct extensive experiments to carefully validate our approach.

\section{Experiments}
\label{sec:experiments}
In this section, we conduct extensive comparisons and analyses to evaluate the proposed FlatMatch method. We first describe the experimental setup and implementation details. Then we compare FlatMatch with many edge-cutting SSL approaches to show the effectiveness of our method. Further, we conduct an ablation study to justify our design of FlatMatch. Finally, we demonstrate the efficiency, stability, and sensitivity of FlatMatch through various analytical studies.

\subsection{Experimental Setup and Details}
We follow the most common semi-supervised image classification protocols by using CIFAR10/100~\cite{krizhevsky2009learning} SVHN~\cite{netzer2011reading}, and STL10~\cite{coates2011analysis} datasets where a various number of labeled data are equally sampled from each class. Following Wang et al.~\cite{wang2023freematch}, We choose Wide ResNet-28-2~\cite{zagoruyko2016wide} for CIFAR10 and SVHN, Wide ResNet-28-8 for CIFAR100, ResNet-37-2~\cite{he2016deep} for STL10. All SSL approaches are implemented using Pytorch framework, and the computation is based on 12GB NVIDIA 3090 GPU. We compare our method with well-known approaches, including $\Pi$-model~\cite{laine2016temporal}, Pseudo Label~\cite{lee2013pseudo}, VAT~\cite{miyato2018virtual}, Mean Teacher~\cite{tarvainen2017mean}, MixMatch~\cite{berthelot2019mixmatch}, ReMixMatch~\cite{berthelot2019remixmatch}, UDA~\cite{xie2020unsupervised}, FixMatch~\cite{sohn2020fixmatch}, Dash~\cite{xu2021dash}, MPL~\cite{pham2020meta}, FlexMatch~\cite{zhang2021flexmatch}, SoftMatch~\cite{chen2023softmatch}, and FreeMatch~\cite{wang2023freematch}.

To optimize all methods, we use SGD with a momentum of $0.9$ with an initial learning rate of $0.03$. We set the batch size as $64$ for all datasets. Moreover, we set the weight decay value, pseudo label threshold $\tau$, unlabeled batch ratio $\mu$, and trade-off for SSL regularization as the same for all compared methods. To implement our FlatMatch, we set the perturbation magnitude $\rho=0.1$, and EMA factor $\alpha=0.999$ which is the same for Mean Teacher and other methods that require EMA. Moreover, in line 9 from Alg.~\ref{alg:flatmatch}, we need to separate the gradients regarding labeled data and unlabeled data in one back-propagation step. Such an operation can be done using ``\data{functorch.make\_functional\_with\_buffers}'' in Pytorch, which can efficiently compute the gradients for each sample. For more details, please see the \textbf{appendix}.

\begin{table}[!t]
	\vspace{-3mm}
	\centering
	\caption{\small Error rates on CIFAR10/100, SVHN, and STL10 datasets. The fully-supervised results of STL10 are unavailable since we do not have label information for its unlabeled data. The best results are highlighted with \textbf{Bold} and the second-best results are highlighted with \underline{underline}.}
	\label{tab:comparsion}
	\resizebox{\textwidth}{!}{%
		\begin{tabular}{l|cccc|ccc|ccc|cc}
			\toprule
			Dataset & \multicolumn{4}{c|}{CIFAR10}& \multicolumn{3}{c|}{CIFAR100}& \multicolumn{3}{c|}{SVHN} & \multicolumn{2}{c}{STL10} \\ \cmidrule(r){1-1}\cmidrule(lr){2-5}\cmidrule(lr){6-8}\cmidrule(lr){9-11}\cmidrule(l){12-13}
			
			\# Label & \multicolumn{1}{c}{10} & \multicolumn{1}{c}{40} & \multicolumn{1}{c}{250}  & \multicolumn{1}{c|}{4000} & \multicolumn{1}{c}{400}  & \multicolumn{1}{c}{2500}  & \multicolumn{1}{c|}{10000} & \multicolumn{1}{c}{40}  & \multicolumn{1}{c}{250}   & \multicolumn{1}{c|}{1000} & \multicolumn{1}{c}{40}  & \multicolumn{1}{c}{1000}\\ \cmidrule(r){1-1}\cmidrule(lr){2-5}\cmidrule(lr){6-8}\cmidrule(lr){9-11}\cmidrule(l){12-13}
			
			$\Pi$-Model~\cite{laine2016temporal} & 79.18{\scriptsize $\pm$1.11} & 74.34{\scriptsize $\pm$1.76} & 46.24{\scriptsize $\pm$1.29} & 13.13{\scriptsize $\pm$0.59} & 86.96{\scriptsize $\pm$0.80} & 58.80{\scriptsize $\pm$0.66} & 36.65{\scriptsize $\pm$0.00} & 67.48{\scriptsize $\pm$0.95} & {13.30\scriptsize $\pm$1.12} & 7.16{\scriptsize $\pm$0.11}  & 74.31{\scriptsize $\pm$0.85} & 32.78{\scriptsize $\pm$0.40} \\
			Pseudo Label~\cite{lee2013pseudo} & 80.21{\scriptsize $\pm$0.55}& 74.61{\scriptsize $\pm$0.26}  & 46.49{\scriptsize $\pm$2.20} & 15.08{\scriptsize $\pm$0.19} & 87.45{\scriptsize $\pm$0.85} & 57.74{\scriptsize $\pm$0.28} & 36.55{\scriptsize $\pm$0.24} & 64.61{\scriptsize $\pm$5.6} & 15.59{\scriptsize $\pm$0.95}  & 9.40{\scriptsize $\pm$0.32}  & 74.68{\scriptsize $\pm$0.99} & 32.64{\scriptsize $\pm$0.71} \\
			VAT~\cite{miyato2018virtual} & 79.81{\scriptsize $\pm$1.17} & 74.66{\scriptsize $\pm$2.12} & 41.03{\scriptsize $\pm$1.79} & 10.51{\scriptsize $\pm$0.12} & 85.20{\scriptsize $\pm$1.40} & 46.84{\scriptsize $\pm$0.79} & 32.14{\scriptsize $\pm$0.19} & 74.75{\scriptsize $\pm$3.38} & 4.33{\scriptsize $\pm$0.12} & 4.11{\scriptsize $\pm$0.20} & 74.74{\scriptsize $\pm$0.38}  & 37.95{\scriptsize $\pm$1.12} \\
			Mean Teacher~\cite{tarvainen2017mean} & 76.37{\scriptsize $\pm$0.44} & 70.09{\scriptsize $\pm$1.60} & 37.46{\scriptsize $\pm$3.30} & 8.10{\scriptsize $\pm$0.21} & 81.11{\scriptsize $\pm$1.44} & 45.17{\scriptsize $\pm$1.06} & 31.75{\scriptsize $\pm$0.23} & 36.09{\scriptsize $\pm$3.98} & 3.45{\scriptsize $\pm$0.03} & 3.27{\scriptsize $\pm$0.05}  & 71.72{\scriptsize $\pm$1.45} & 33.90{\scriptsize $\pm$1.37} \\
			MixMatch~\cite{berthelot2019mixmatch} & 65.76{\scriptsize $\pm$7.06} & 36.19{\scriptsize $\pm$6.48} & 13.63{\scriptsize $\pm$0.59} & 6.66{\scriptsize $\pm$0.26} & 67.59{\scriptsize $\pm$0.66} & 39.76{\scriptsize $\pm$0.48} & 27.78{\scriptsize $\pm$0.29} & 30.60{\scriptsize $\pm$8.39} & 4.56{\scriptsize $\pm$0.32} & 3.69{\scriptsize $\pm$0.37}  & 54.93{\scriptsize $\pm$0.96} & 21.70{\scriptsize $\pm$0.68} \\
			ReMixMatch~\cite{berthelot2019remixmatch} & 20.77{\scriptsize $\pm$7.48} & 9.88{\scriptsize $\pm$1.03} & 6.30{\scriptsize $\pm$0.05} & 4.84{\scriptsize $\pm$0.01} & 42.75{\scriptsize $\pm$1.05} & 26.03{\scriptsize $\pm$0.35} &  20.02{\scriptsize $\pm$0.27} & 24.04{\scriptsize $\pm$9.13} & 6.36{\scriptsize $\pm$0.22}  & 5.16{\scriptsize $\pm$0.31} & 32.12{\scriptsize $\pm$6.24} & 6.74{\scriptsize $\pm$0.14} \\
			UDA~\cite{xie2020unsupervised} & 34.53{\scriptsize $\pm$10.69} & 10.62{\scriptsize $\pm$3.75} & 5.16{\scriptsize $\pm$0.06} & 4.29{\scriptsize $\pm$0.07} & 46.39{\scriptsize $\pm$1.59} & 27.73{\scriptsize $\pm$0.21} & 22.49{\scriptsize $\pm$0.23} & 5.12{\scriptsize $\pm$4.27} & 1.92{\scriptsize $\pm$0.05} & 1.89{\scriptsize $\pm$0.01}  & 37.42{\scriptsize $\pm$8.44} & 6.64{\scriptsize $\pm$0.17} \\
			FixMatch~\cite{sohn2020fixmatch} & 24.79{\scriptsize $\pm$7.65} & 7.47{\scriptsize $\pm$0.28} & 4.86{\scriptsize $\pm$0.05} & 4.21{\scriptsize $\pm$0.08} & 46.42{\scriptsize $\pm$0.82} & 28.03{\scriptsize $\pm$0.16} & 22.20{\scriptsize $\pm$0.12} & 3.81{\scriptsize $\pm$1.18} & 2.02{\scriptsize $\pm$0.02}  & 1.96{\scriptsize $\pm$0.03} & 35.97{\scriptsize $\pm$4.14} & 6.25{\scriptsize $\pm$0.33} \\
			Dash~\cite{xu2021dash} & 27.28{\scriptsize $\pm$14.09} & 8.93{\scriptsize $\pm$3.11} & 5.16{\scriptsize $\pm$0.23} & 4.36{\scriptsize $\pm$0.11} & 44.82{\scriptsize $\pm$0.96} & 27.15{\scriptsize $\pm$0.22} & 21.88{\scriptsize $\pm$0.07} & 2.19{\scriptsize $\pm$0.18} & 2.04{\scriptsize $\pm$0.02} & 1.97{\scriptsize $\pm$0.01}  & 34.52{\scriptsize $\pm$4.30} & 6.39{\scriptsize $\pm$0.56} \\
			MPL~\cite{pham2020meta} & 23.55{\scriptsize $\pm$6.01} & 6.62{\scriptsize $\pm$0.91} & 5.76{\scriptsize $\pm$0.24} & 4.55{\scriptsize $\pm$0.04} & 46.26{\scriptsize $\pm$1.84} & {27.71\scriptsize $\pm$0.19} & 21.74{\scriptsize $\pm$0.09} & 9.33{\scriptsize $\pm$8.02} & 2.29{\scriptsize $\pm$0.04} & 2.28{\scriptsize $\pm$0.02}  & 35.76{\scriptsize $\pm$4.83} & 6.66{\scriptsize $\pm$0.00} \\
			FlexMatch~\cite{zhang2021flexmatch} & 13.85{\scriptsize $\pm$12.04} & 4.97{\scriptsize $\pm$0.06} & 4.98{\scriptsize $\pm$0.09} & 4.19{\scriptsize $\pm$0.01} & 39.94{\scriptsize $\pm$1.62} & 26.49{\scriptsize $\pm$0.20} & 21.90{\scriptsize $\pm$0.15} & 8.19{\scriptsize $\pm$3.20} & 6.59{\scriptsize $\pm$2.29} & 6.72{\scriptsize $\pm$0.30} & 29.15{\scriptsize $\pm$4.16} & 5.77{\scriptsize $\pm$0.18} \\
			SoftMatch~\cite{chen2023softmatch} & - & 4.91{\scriptsize $\pm$0.12} & 4.82{\scriptsize $\pm$0.09} & 4.04{\scriptsize $\pm$0.02} & \underline{37.10}{\scriptsize $\pm$0.77} & 26.66{\scriptsize $\pm$0.25} & 22.03{\scriptsize $\pm$0.03} & 2.33{\scriptsize $\pm$0.25} & - & 2.01{\scriptsize $\pm$0.01} & 21.42{\scriptsize $\pm$3.48} & 5.73{\scriptsize $\pm$0.24} \\
			FreeMatch~\cite{wang2023freematch} & \underline{8.07}{\scriptsize $\pm$4.24} & \underline{4.90}{\scriptsize $\pm$0.04} & 4.88{\scriptsize $\pm$0.18} & 4.10{\scriptsize $\pm$0.02} & 37.98{\scriptsize $\pm$0.42} & 26.47{\scriptsize $\pm$0.20} & 21.68{\scriptsize $\pm$0.03} & \underline{1.97}{\scriptsize $\pm$0.02} & 1.97{\scriptsize $\pm$0.01} & 1.96{\scriptsize $\pm$0.03} & \underline{15.56}{\scriptsize $\pm$0.55} & 5.63{\scriptsize $\pm$0.15} \\
			FlatMatch & 15.23{\scriptsize $\pm$8.67} & 5.58{\scriptsize $\pm$2.36} & \textbf{4.22}{\scriptsize $\pm$1.14} & \textbf{3.61}{\scriptsize $\pm$0.49} & 38.76{\scriptsize $\pm$1.62} & \textbf{25.38}{\scriptsize $\pm$0.85} & \textbf{19.01}{\scriptsize $\pm$0.43} & 2.46{\scriptsize $\pm$0.06} & \textbf{1.43}{\scriptsize $\pm$0.05} & \textbf{1.41}{\scriptsize $\pm$0.04} & 16.20{\scriptsize $\pm$4.34} & \textbf{4.82}{\scriptsize $\pm$1.21} \\
			\midrule
			FlatMatch-e & 15.69{\scriptsize $\pm$6.35} & 5.63{\scriptsize $\pm$1.87} & \underline{4.53}{\scriptsize $\pm$1.85} & \underline{3.57}{\scriptsize $\pm$0.50} & 38.98{\scriptsize $\pm$1.53} & \underline{25.62}{\scriptsize $\pm$0.88} & \underline{19.78}{\scriptsize $\pm$0.89} & 2.66{\scriptsize $\pm$0.09} & \underline{1.47}{\scriptsize $\pm$0.08} & \underline{1.46}{\scriptsize $\pm$0.07} & 16.32{\scriptsize $\pm$4.64} & \underline{5.03}{\scriptsize $\pm$1.06} \\
			FlatMatch (\scriptsize{Fix label}) & \textbf{7.36}{\scriptsize $\pm$5.62} & \textbf{4.89}{\scriptsize $\pm$1.24} & \textbf{3.90}{\scriptsize $\pm$1.72} & \textbf{3.61}{\scriptsize $\pm$0.49} & \textbf{36.97}{\scriptsize $\pm$0.95} & \textbf{25.38}{\scriptsize $\pm$0.85} & \textbf{19.01}{\scriptsize $\pm$0.43} & \textbf{2.14}{\scriptsize $\pm$0.05} & \textbf{1.43}{\scriptsize $\pm$0.05} & \textbf{1.41}{\scriptsize $\pm$0.04} & \textbf{14.96}{\scriptsize $\pm$0.67} & \textbf{4.82}{\scriptsize $\pm$1.21} \\
			\midrule
			Fully-Supervised    & \multicolumn{4}{c|}{4.62{\scriptsize $\pm$0.05}} & \multicolumn{3}{c|}{19.30{\scriptsize $\pm$0.09}}  & \multicolumn{3}{c|}{2.13{\scriptsize $\pm$0.01}} & \multicolumn{2}{c}{-}\\
			\bottomrule
		\end{tabular}
	}
	\vspace{-0.2in}
\end{table}

\subsection{Quantitative Comparison}
The main comparison results are shown in Table~\ref{tab:comparsion}, we can see that our FlatMatch achieves state-of-the-art results on many scenarios. For example, on CIFAR100 with 2500 labels, FlatMatch achieves 25.38\% test errors, surpassing the second-best method, FreeMatch, for 1.09\%; and on CIFAR100 with 10000 labels, FlatMatch achieves 19.01\% performance, further improving the second-best result, ReMixMatch, for 1.01\%, which even passes the fully-supervised baseline result for 0.29\%. Moreover, FlatMatch reaches extremely low error rates in many settings. For instance, in CIFAR10 with 250 labels and 4000 labels, FlatMatch only has 4.22\% and 3.61\% error rates; and in SVHN with 250 labels and 1000 labels, FlatMatch can also surpass the fully-supervised learning by producing 1.43\% and 1.41\% error rates, respectively. Additionally, FlatMatch breaks the record on the STL10 dataset with 1000 labels by reaching 4.82\% performance.

\begin{wrapfigure}{r}{5.2cm}
	\vspace*{-3.5mm}
	\centering
	\includegraphics[width=\linewidth]{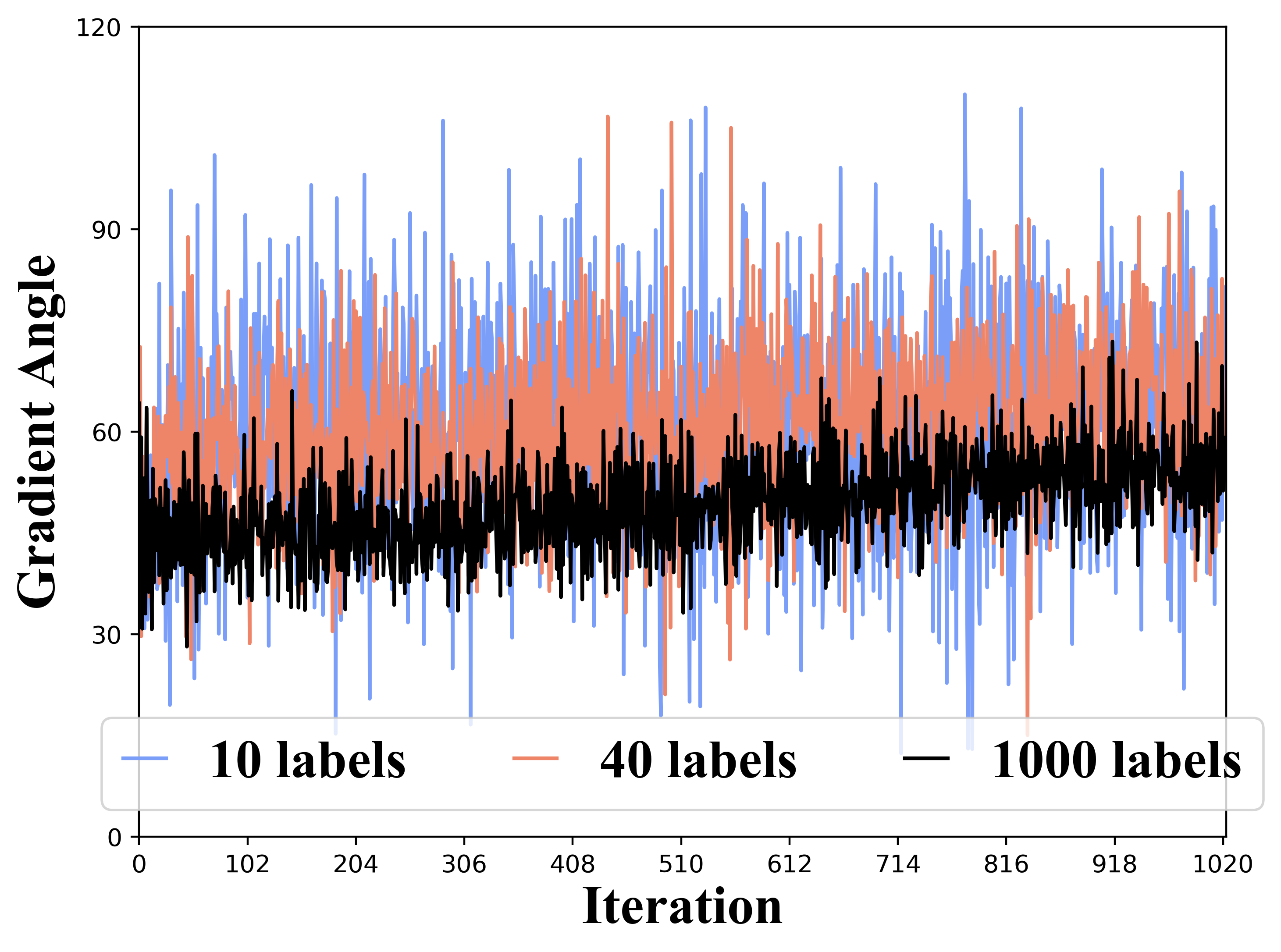}
	\caption{\small Gradient angle between cross-entropy loss $\mathcal{L}_l$ and general SSL loss $\mathcal{L}_{ssl}$.}
	\label{fig:angle}
	\vspace*{-3.5mm}
\end{wrapfigure}
Moreover, we also show the performance of FlatMatch-e, an efficient variant of FlatMatch, which can also surpass many existing state-of-the-art methods. Such as in CIFAR100 with 10000 labels, FlatMatch-e can beat the second-best method, SoftMatch, with a 2.04\% improvement. More importantly, FlatMatch-e only shows slight performance drops from FlatMatch but significantly reduces the computational burden, which is demonstrated in Section~\ref{sec:study}.

Despite its effective performance, we also noticed that in some scenarios where the labels are extremely limited, such as CIFAR10 with 10 labels, CIFAR100 with 400 labels, etc, our FlatMatch is slightly worse than a recently proposed method FreeMatch~\cite{wang2023freematch}. This is because the gradient estimated by the extremely scarce labeled data cannot align with the general training direction. To intuitively demonstrate this issue, we show the gradient angle between the cross-entropy loss from labeled data and the total SSL loss in Fig.~\ref{fig:angle}. We can see that the gradient angle oscillates stronger as the number of labels gets smaller. When the label number reduces to only 10, the angle becomes extremely unstable and could surpass 90$^{\circ}$. As a result, such a phenomenon could affect the worst-case perturbation computed by FlatMatch, which would introduce negligible noise to the training process.

\renewcommand{\algorithmicrequire}{\textbf{Input:}}
\renewcommand{\algorithmicensure}{\textbf{Output:}}
\begin{algorithm}[t]
	\small
	\caption{\small FlatMatch with fixed labels for Stabilizing Cross-Sharpness}
	\label{alg:fixlabel}
	\begin{algorithmic}[1]
		\Require Labeled dataset $\mathcal{D}^l$, unlabeled dataset $\mathcal{D}^u$, model $\theta$, number of fixed labels $\#fix$, number of pre-train epochs $\#pre\_train$.
		\For{$t \in 0,1, \ldots, T-1$}
		\If{{$t \in 0,1, \ldots, \#pre\_train$}}
		\State SSL pre-training by optimizing Eq.~\ref{eq:semi_obj};
		\If{$t=\#pre\_train$}
		\State Select $\text{Top-}{\#fix}$ confident unlabeled data $x^u_i$ and assign fixed labels $\hat{y}_i:=\text{argmax}(g_{\theta}(x^u_i))$;
		\EndIf
		\Else
		\State Using the augmented labeled data to compute cross-sharpness via Eq.~\ref{eq:cross_sharpness};
		\State Using the original $\mathcal{D}^l$ and $\mathcal{D}^u$ to optimize $\theta$ via Eq.~\ref{eq:optimization};
		\EndIf
		\EndFor
	\end{algorithmic}
\end{algorithm}

However, this drawback can be properly solved by slightly augmenting the number of labels with pseudo-labeled unlabeled data. Specifically, before we start minimizing cross-sharpness, we first pre-train the backbone model with a common SSL method, such as FixMatch or FreeMatch, for 16 epochs to increase the confidence on unlabeled data. Then, we choose the most confident, \textit{i.e.} high softmax probability, unlabeled data to be pseudo-labeled as labeled data. Different from other unlabeled data that are also trained with pseudo labels, the selected unlabeled data are fixed with their labels and act as labeled data when computing the sharpness. In this way, the cross-sharpness would be computed accurately by using both the original labeled data and augmented labeled data. Note that the augmentation with fixed labels is \textbf{only} used in the sharpness computation, when conducitng the second propagation step, all unlabeled data are treated similar to general SSL strategy, such process is summarized in Algorithm~\ref{alg:fixlabel}. Under this training strategy, we dub our method as ``FlatMatch (Fix label)'' and show its result in Table~\ref{tab:comparsion}. The number of fixed labeled data in CIFAR10, CIFAR100, SVHM, and STL10 is set to 500, 2500, 500, and 1000, respectively. If the number of the original labeled dataset contains enough labels, we do not add more fixed labeled data. We can see that in this scenario, our method has the best performance in all settings including the ones with 1 or 4 labels in each class, which demonstrates that using more labels can largely stabilize sharpness computation and further benefits the performance of FlatMatch, thus achieving superior results on all settings.

\vspace*{-3mm}
\subsection{Ablation Study}
\vspace*{-2mm}
To carefully justify the design of our method, we compare FlatMatch with ``sharp. on labeled data $\mathcal{D}^l$'' and ``sharp. on unlabeled data $\mathcal{D}^u$'' where the former one denotes both the worst-case model $\hat{\theta}$ and sharpness is computed on labeled data and the latter one denotes $\hat{\theta}$ and sharpness are calculated on unlabeled data. Additionally, we compute the sharpness on the full dataset, as denoted by ``sharp. on unlabeled data $\mathcal{D}^l\cup\mathcal{D}^u$''. Moreover, we also analyze the effect of EMA smoothing on the performance of FlatMatch-e. Specifically, we compare FlatMatch-e with setting ``w/o EMA'' that just uses a gradient from the last batch of labeled data to calculate our worst-case model. The ablation study is conducted using CIFAR100 with a varied number of labels, which is shown in Table~\ref{tab:ablation}. 

\begin{wraptable}{r}{6.8cm}
	\vspace*{-3.5mm}
	\centering
	\small
	\caption{\small Ablation study on CIFAR100.}
	\label{tab:ablation}
	\setlength{\tabcolsep}{0.6mm}
	\begin{tabular}{l|ccc}
		\toprule
		Dataset & \multicolumn{3}{c}{CIFAR100} \\ \cmidrule(r){1-1}\cmidrule(l){2-4}
		
		\# Label & \multicolumn{1}{c}{400}  & \multicolumn{1}{c}{2500}  & \multicolumn{1}{c}{10000}\\ \cmidrule(r){1-1}\cmidrule(l){2-4}
		
		sharp. on $\mathcal{D}^l$ & 42.63{\scriptsize $\pm$0.34} & 26.85{\scriptsize $\pm$0.45} & 21.79{\scriptsize $\pm$0.24} \\
		sharp. on $\mathcal{D}^u$ & 49.45{\scriptsize $\pm$2.76} & 36.30{\scriptsize $\pm$2.01} & 27.05{\scriptsize $\pm$2.98} \\
		
		sharp. on $\mathcal{D}^l\cup\mathcal{D}^u$ & 43.88{\scriptsize $\pm$1.64} & 27.26{\scriptsize $\pm$1.62} & 23.42{\scriptsize $\pm$1.77} \\
		FlatMatch & \textbf{38.76}{\scriptsize $\pm$1.62} & \textbf{25.38}{\scriptsize $\pm$0.85} & \textbf{19.01}{\scriptsize $\pm$0.43} \\
		w/o EMA & 40.64{\scriptsize $\pm$0.97} & 29.44{\scriptsize $\pm$1.56} & 23.23{\scriptsize $\pm$1.28} \\
		FlatMatch-e & \underline{38.98}{\scriptsize $\pm$1.53} & \underline{25.62}{\scriptsize $\pm$0.88} & \underline{19.78}{\scriptsize $\pm$0.89} \\
		\bottomrule
	\end{tabular}
	\vspace*{-3.5mm}
\end{wraptable}
First, we can see that both two choices of our method are effective. Particularly, computing sharpness only on labeled data $\mathcal{D}^l$ shows smaller performance degradation than computing sharpness on unlabeled data $\mathcal{D}^u$, and it even shows better results than FixMatch. Hence, we know that the sharpness of labeled data can indeed improve the performance of SSL, only having limited performance because the number of labeled data is too scarce. On the other hand, when sharpness is completely based on unlabeled data, the performance significantly drops by nearly 10\% compared to ``sharpness on $\mathcal{D}^l$''. This is because the training process on unlabeled data contains too much noise which causes erroneous gradient computation that would hinder the effectiveness of penalizing sharpness. Furthermore, we find that ``w/o EMA'' shows slightly inferior performance to FlatMatch-e. Such degradation is consistent with the findings from Liu et al.~\cite{liu2022towards} that the last gradient direction is distinct from the current one. As gradient descent has been conducted in the previous batch, reusing the gradient to perturb the current model might not find the perfect worst-case model on the current labeled data. Based on our observation, using EMA can stabilize the gradient can lead to accurate sharpness calculation.

\begin{figure*}[h]
	\begin{minipage}[t]{0.32\textwidth}
		\centering
		\text{\small \ 2D contours of labeled data}
		\includegraphics[width=\linewidth]{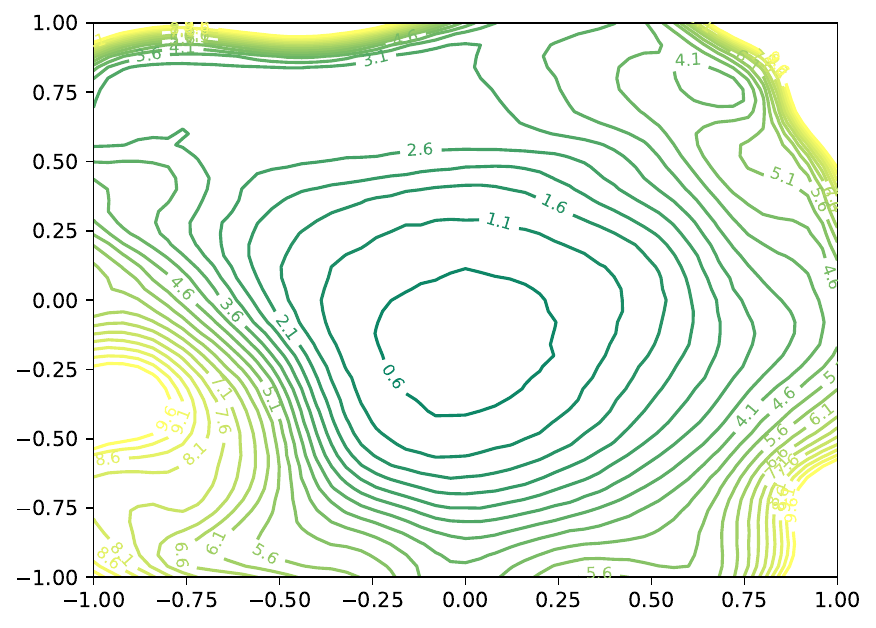}
	\end{minipage}
	\begin{minipage}[t]{0.32\textwidth}
		\centering
		\text{\small \ 2D contours of unlabeled data}
		\includegraphics[width=\linewidth]{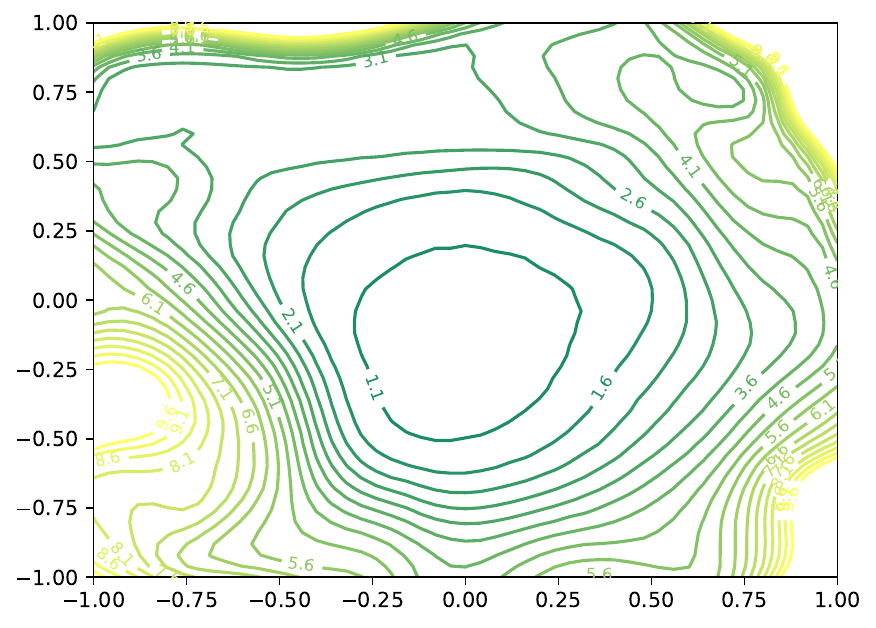}
	\end{minipage}
	\begin{minipage}[t]{0.35\textwidth}
		\centering
		\text{\small1D loss curves}
		\raisebox{0.01\height}{\includegraphics[width=\linewidth]{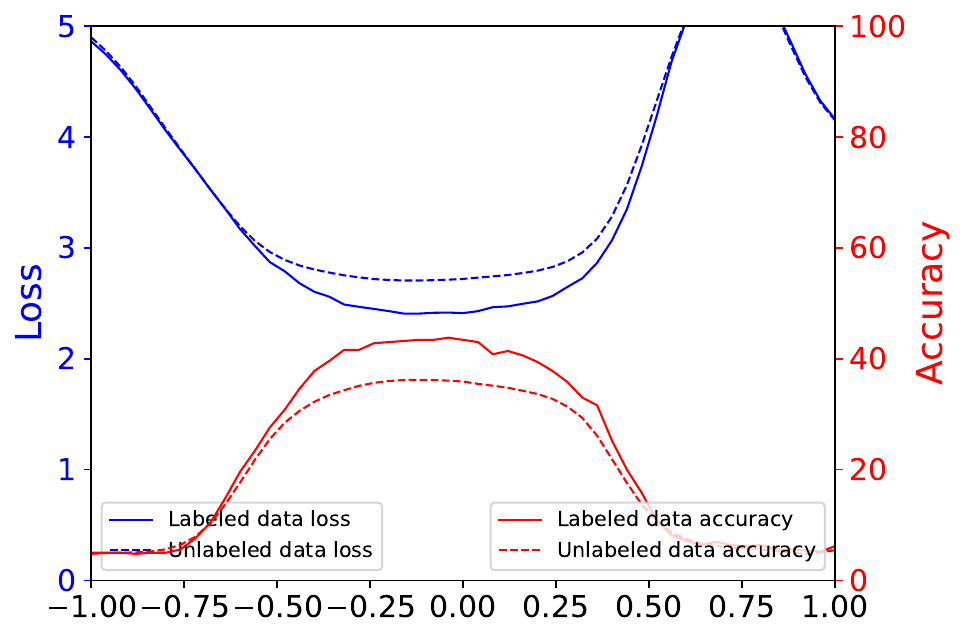}}
	\end{minipage}
	\begin{minipage}[t]{0.32\textwidth}
		\centering
		\includegraphics[width=\linewidth]{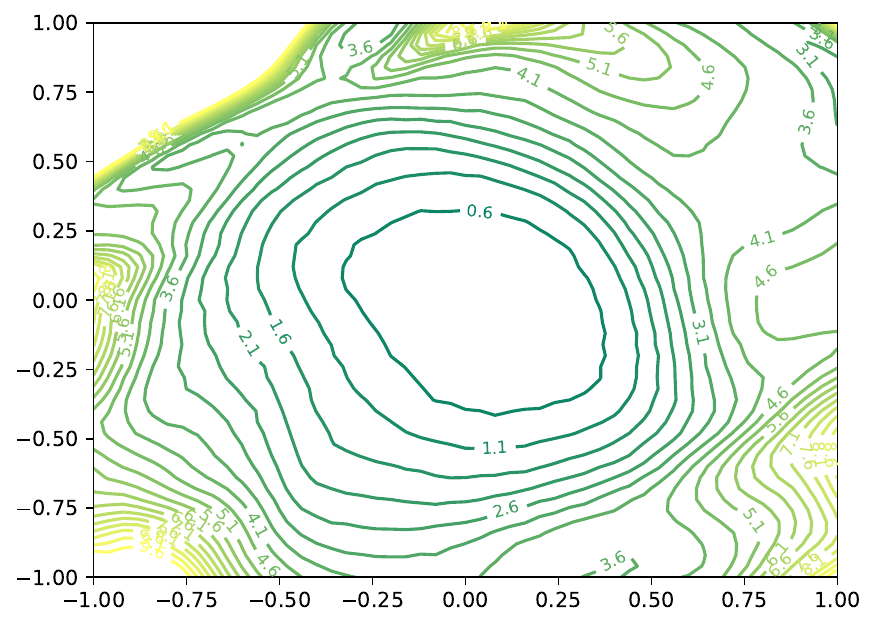}
	\end{minipage}
	\begin{minipage}[t]{0.32\textwidth}
		\centering
		\includegraphics[width=\linewidth]{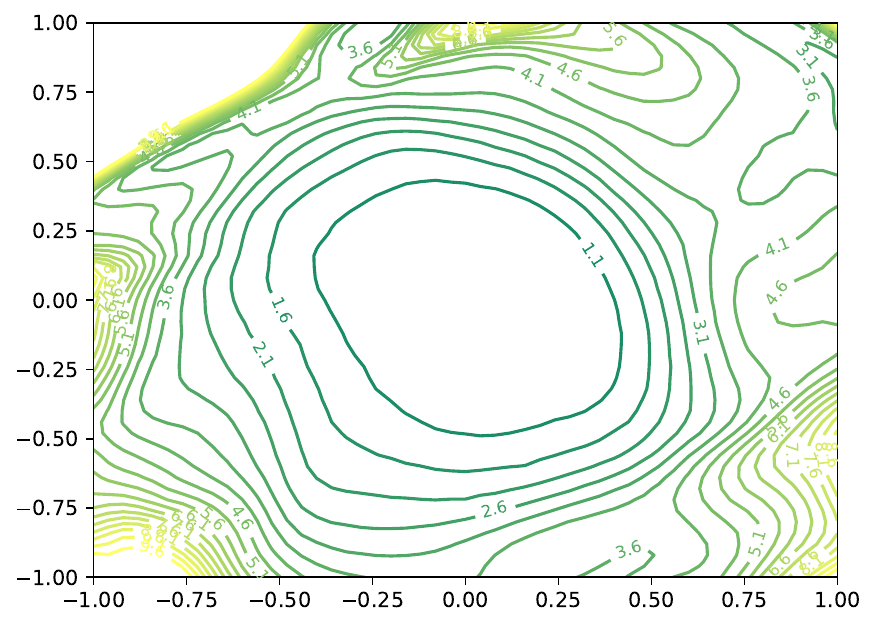}
	\end{minipage}
	\begin{minipage}[t]{0.35\textwidth}
		\centering
		\includegraphics[width=\linewidth]{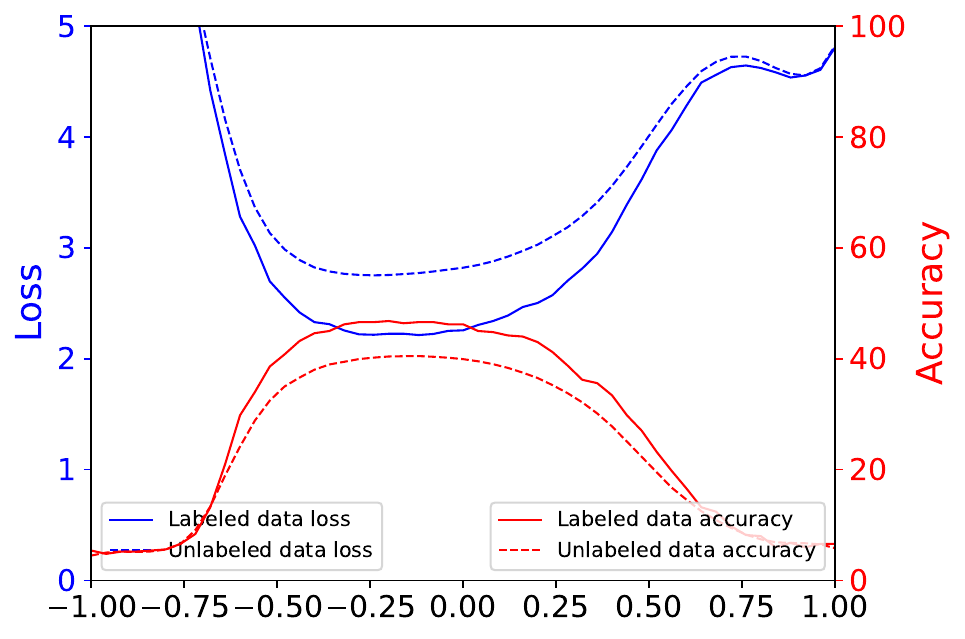}
	\end{minipage}
	\caption{\small Loss landscapes of labeled data and unlabeled data obtained simultaneously from training using FlatMatch on CIFAR10 with 250 labels per class. The first row and second row show the results obtained from epoch 60 and epoch 150, respectively. The first column and second column show the 2D loss contours of labeled data and unlabeled data, respectively, and the last column shows the 1D loss curves.}
	\label{fig:visualization}
\end{figure*}

\subsection{Analytical Study}
\label{sec:study}
In this section, we analyze the performance of FlatMatch by considering visualization, parameter sensitivity, training stability, and .efficiency.

\textbf{Loss visualization:} To show that FlatMatch can properly solve the sharp loss problem of labeled data, we train FlatMatch under the same setting as the ones demonstrated in Section~\ref{sec:introduction} and plot the 2D loss contour as well as the 1D loss curve in Fig.~\ref{fig:visualization}. We can see that our FlatMatch can produce a much flatter loss landscape than FixMatch does in Fig.~\ref{fig:motivation}, where the jagged curve has been eliminated and become very smooth. Therefore, by minimizing cross-sharpness, the generalization performance on labeled data can be largely improved.

\begin{figure*}[t]
	\vspace*{-3mm}
	\begin{minipage}[t]{0.304\textwidth}
		\centering
		\text{\small \ Parameter Sensitivity}
		\includegraphics[width=\linewidth]{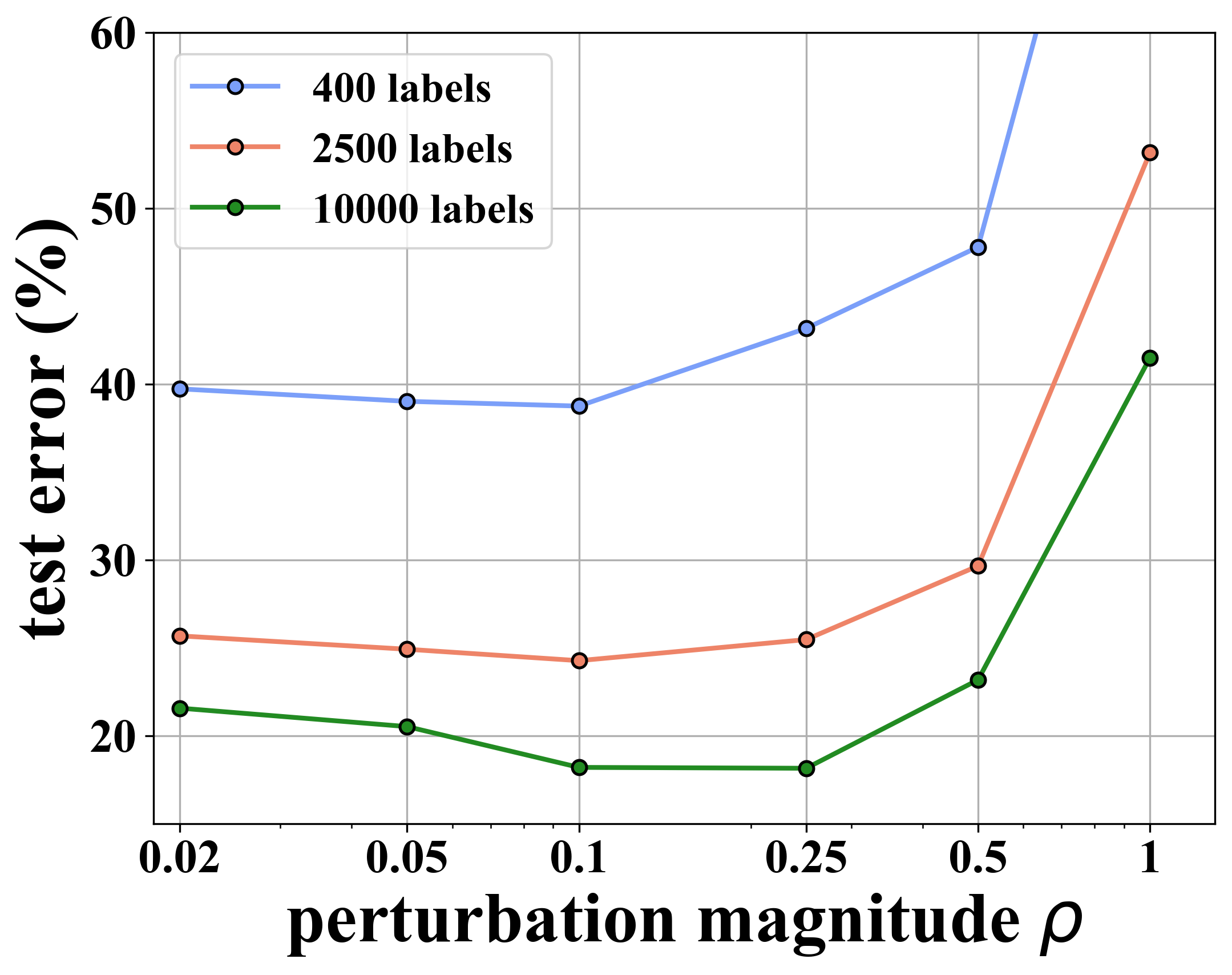}
		\caption{\small Parameter sensitivity analysis regarding perturbation magnitude $\rho$ on CIFAR100.}
		\label{fig:sensitivity}
	\end{minipage}
	\ \ 
	\begin{minipage}[t]{0.304\textwidth}
		\centering
		\text{\small \ Training Stability}
		\includegraphics[width=\linewidth]{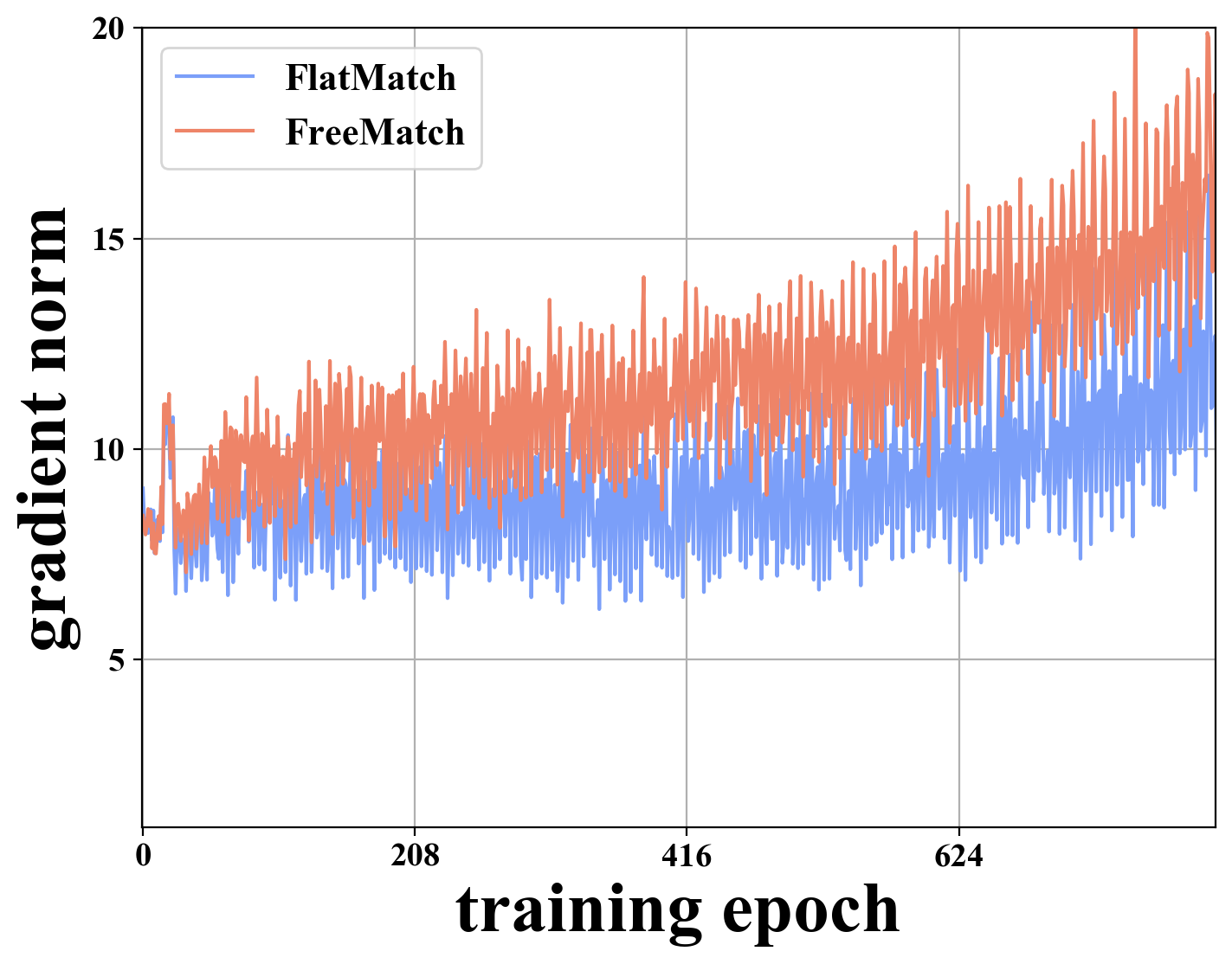}
		\caption{\small Stability analysis on gradient norm from training on CIFAR100.}
		\label{fig:gradnorm}
	\end{minipage}
	\ \ 
	\begin{minipage}[t]{0.352\textwidth}
		\centering
		\text{\small\ Training Efficiency}
		\includegraphics[width=\linewidth]{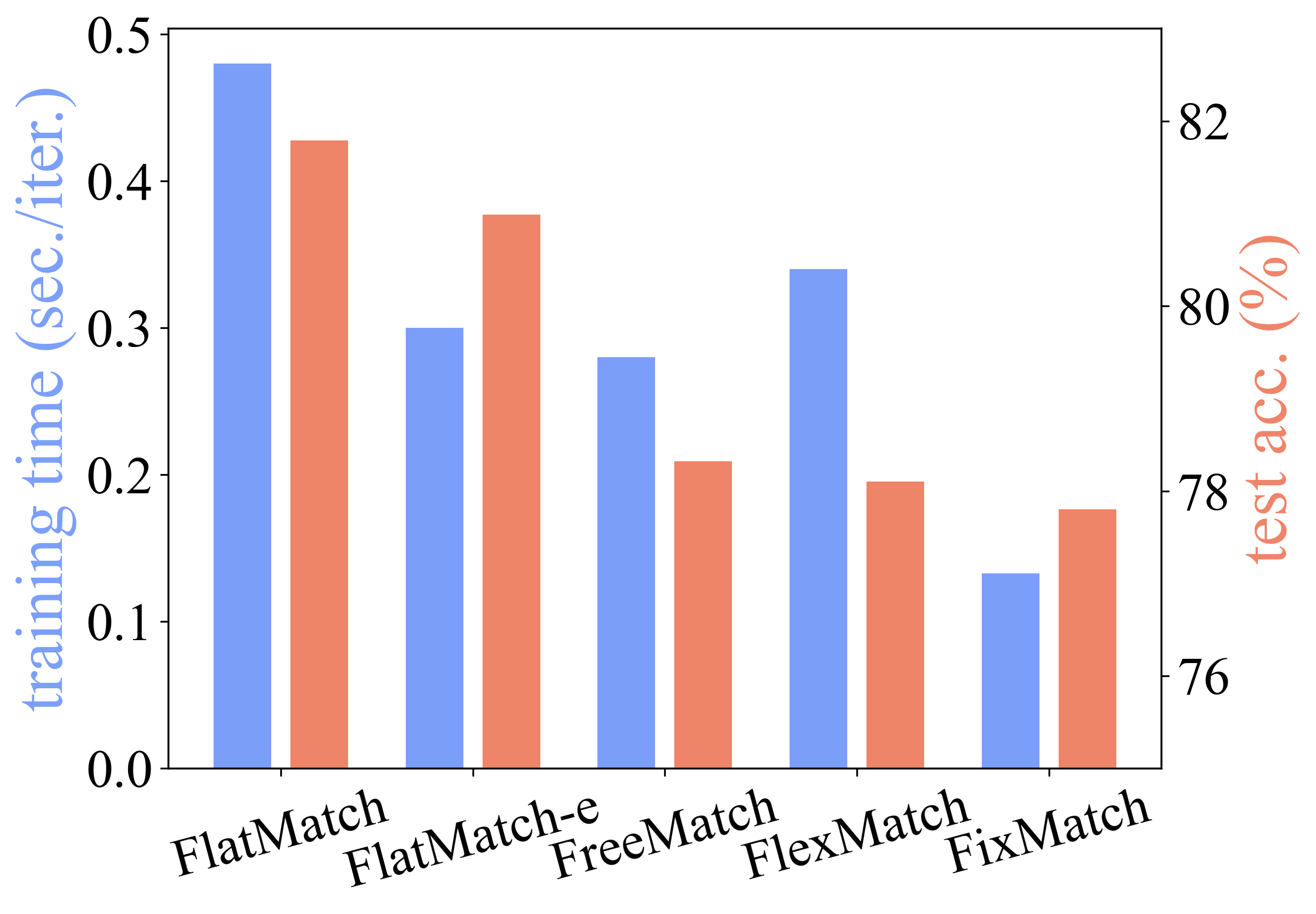}
		\caption{\small Training efficiency comparison on CIFAR100 with 10000 labels.}
		\label{fig:efficiency}
	\end{minipage}
	\vspace*{-3mm}
\end{figure*}

\textbf{Sensitivity analysis:} Our FlatMatch requires a hyperparameter $\rho$ which controls the perturbation magnitude to obtain the worst-case model. To analyze the performance sensitivity of varying $\rho$, we show the result on CIFAR100 in Fig.~\ref{fig:sensitivity}. We observe that small $\rho$ values show little impact on the test performance. However, when $\rho$ increases to more than 0.25, the performance would largely degrade. Moreover, among three settings with varied label numbers, we find that more numbers labels could enhance the model sensitivity against changing of $\rho$. For example, when changing the $\rho$ from 0.1 to 0.25, the performance difference on 400 labels, 2500 labels, and 10000 labels are 0.05\%, 1.2\%, 3.42\%. Generally, the optimal value for $\rho$ is 0.1.

\textbf{Training stability:} To validate the training stability of FlatMatch, we show the gradient norm which is an important criterion to present gradient variation in Fig.~\ref{fig:gradnorm}. We observe that the gradient norm of FlatMatch is significantly smaller than FreeMatch during the whole training phase. Therefore, minimizing the cross-sharpness can indeed improve training stability.

\textbf{Training efficiency:} We compare the training time and test accuracy of FlatMatch, FlatMatch-e, FreeMatch, FlexMatch, and FixMatch to validate the efficiency property of our method. As shown in Fig.~\ref{fig:efficiency}, we find that despite the superior accuracy performance, FlatMatch requires more time for each iteration than the other three methods. However, the efficient variant FlatMatch-e can largely reduce the computational cost without losing too much learning performance. Hence, we can conclude that leveraging EMA for computing cross-sharpness is both effective and efficient.

\section{Conclusion, Limitation and Broader Impact}
\label{sec:conclusion}
In this paper, we propose a novel SSL method dubbed FlatMatch to address the mismatched generalization performance between labeled data and unlabeled data. By minimizing the proposed cross-sharpness regularization, FlatMatch can leverage the richness of unlabeled data to improve the generalization performance on labeled data. As a result, the supervised knowledge from labeled data can better guide SSL, and achieve improved test performance than most existing methods. Moreover, we propose an efficient implementation to reduce the computation cost. We conduct comprehensive experiments to validate our method regarding effectiveness, sensitivity, stability, and efficiency. Additionally, FlatMatch is slightly limited under barely-supervised learning due to the requirement of enough labeled data. Our method shows that SSL can be further improved by exploring generalization, which could be a potential direction in future research.

\section{Acknowledgements}
Li Shen was partially supported by STI 2030—Major Projects (2021ZD0201405). Zhuo Huang was supported by JD Technology Scholarship for Postgraduate Research in Artificial Intelligence (SC4103). Jun Yu was supported by the Natural Science Foundation of China (62276242), National Aviation Science Foundation (2022Z071078001), CAAI-Huawei MindSpore Open Fund (CAAIXSJLJJ-2021-016B, CAAIXSJLJJ-2022-001A), Anhui Province Key Research and Development Program (202104a05020007), USTC-IAT Application Sci. \& Tech. Achievement Cultivation Program (JL06521001Y), and Sci. \& Tech. Innovation Special Zone (20-163-14-LZ-001-004-01). Bo Han was supported by the NSFC Young Scientists Fund (62006202), NSFC General Program (62376235), and Guangdong Basic and Applied Basic Research Foundation (2022A1515011652). Tongliang Liu was partially supported by the following Australian Research Council projects: FT220100318, DP220102121, LP220100527, LP220200949, and IC190100031.

{
\bibliographystyle{ieee}
\bibliography{egbib}
}

\clearpage
\begin{center}
	\vspace{5mm}
	\rule{\linewidth}{3pt}\\ 
	{\Large\bf Supplementary Material for\\ ``FlatMatch: Bridging Labeled Data and Unlabeled Data with Cross-Sharpness for Semi-Supervised Learning''}
	\rule{\linewidth}{1pt}
\end{center}
\vspace{6mm}
\appendix

In this Appendix, we provide additional details and experimental results to complement the proposed method. First, we describe supplementary experimental details in Section~\ref{sec:supp_details}. Then, we provide extra quantitative results, including employing FlatMatch to other SSL methods, comparisons on ImageNet30 and ImageNet~\cite{russakovsky2015imagenet} datasets, performance on Vision Transformer in Section~\ref{sec:quantitative}. Further, we show more empirical results to qualitatively validate FlatMatch in Section~\ref{sec:qualitative}. Then, we conduct convergence study on test accuracy and training loss curves in Section~\ref{sec:convergence}. Moreover, we provide extra visualizations to show the loss landscape on later state of training in Section~\ref{sec:more_vis}. Finally, we summarize this paper and make a discussion on prospective research in Section~\ref{sec:summary}.

\section{Supplementary Details}\label{sec:supp_details}
The experimental setting of this paper follows Wang et al.~\cite{wang2022usb}. Specifically, the hyper-parameters are composed of algorithm-dependent parameters and algorithm-independent parameters, which are shown in Table~\ref{tab:alg_dep} and Table~\ref{tab:alg_ind}, respectively. For algorithm-dependent parameters of FlatMatch, we use the same unlabeled data and labeled data ratio as FreeMatch~\cite{wang2023freematch} as well as all other baseline methods to sample data into a mini-batch. The perturbation magnitude $\alpha$ is based on the results from hyper-parameter sensitivity analysis in the main paper and is chosen as 0.05 for all experiments. For updating the historical gradient using a memory buffer, we use EMA with factor $\alpha$ to ensemble the gradient result. Moreover, we choose the thresholding strategy from FreeMatch and use an EMA decay. Note that for combining the cross-sharpness regularization from FlatMatch with empirical risk, we find that there is no need to introduce another weight to trade off the two loss functions, hence the weight for cross-sharpness is just set to 1 for all experiments. For algorithm-independent hyper-parameters, we have listed the important model setting, optimizer parameters, and data sampling setting as below. Note that all baseline methods follow the implementation of USB~\cite{wang2022usb} and are trained with EMA decay with 0.999 to smooth the parameter updating.

\begin{table}[!htbp]
	\centering
	\caption{Algorithm-dependent hyper-parameters.}
	\label{tab:alg_dep}
	\begin{adjustbox}{width=0.9\columnwidth, center}
		\begin{tabular}{cccccc}
			\toprule
			Algorithm &  FlatMatch \\\cmidrule(r){1-1} \cmidrule(lr){2-2}
			Unlabeled Data to Labeled Data Ratio (CIFAR-10/100, STL-10, SVHN)    &  7 \\
			\cmidrule(r){1-1}\cmidrule(lr){2-2}
			Unlabeled Data to Labeled Data Ratio (ImageNet30)    &  1 \\
			\cmidrule(r){1-1}\cmidrule(lr){2-2}
			Perturbation magnitude $\rho$ for all experiments    &  0.05 \\
			\cmidrule(r){1-1}\cmidrule(lr){2-2}
			EMA factor $\alpha$ for updating gradient   &  0.999 \\
			\cmidrule(r){1-1}\cmidrule(lr){2-2}
			Thresholding EMA decay for all experiments  & 0.999 \\
			\cmidrule(r){1-1}\cmidrule(lr){2-2}
			Trade-off weight $\lambda_{\text{X-sharp}}$ for cross-sharpness  &  1 \\
			\bottomrule
		\end{tabular}
	\end{adjustbox}
\end{table}

\begin{table}[!htbp]
	\centering
	\caption{Algorithm-independent hyper-parameters.}
	\label{tab:alg_ind}
	\begin{adjustbox}{width=0.9\columnwidth, center}
		\begin{tabular}{cccccc}\toprule
			Dataset &  CIFAR-10 & CIFAR-100 & STL-10 & SVHN & ImageNet30 \\\cmidrule(r){1-1} \cmidrule(lr){2-2}\cmidrule(lr){3-3}\cmidrule(lr){4-4}\cmidrule(lr){5-5}\cmidrule(l){6-6}
			Model    &  WRN-28-2 & WRN-28-8 & WRN-37-2 & WRN-28-2 & ResNet-50 \\\cmidrule(r){1-1} \cmidrule(lr){2-2}\cmidrule(lr){3-3}\cmidrule(lr){4-4}\cmidrule(lr){5-5}\cmidrule(l){6-6}
			Weight decay&  5e-4  & 1e-3 & 5e-4 & 5e-4 & 3e-4\\ \cmidrule(r){1-1} \cmidrule(lr){2-2}\cmidrule(lr){3-3}\cmidrule(lr){4-4}\cmidrule(lr){5-5}\cmidrule(l){6-6}
			Batch size & \multicolumn{4}{c}{64} & \multicolumn{1}{c}{128}\\\cmidrule(r){1-1} \cmidrule(lr){2-5} \cmidrule(l){6-6}
			Learning rate & \multicolumn{5}{c}{0.03}\\\cmidrule(r){1-1} \cmidrule(l){2-6}
			SGD momentum & \multicolumn{5}{c}{0.9}\\\cmidrule(r){1-1} \cmidrule(l){2-6}
			EMA decay & \multicolumn{5}{c}{0.999}\\
			\bottomrule
		\end{tabular}
	\end{adjustbox}
\end{table}

\begin{table}[!htbp]
	\centering
	\caption{Performance on boosting other SSL methods using FlatMatch with the fixed number of labels.}
	\label{tab:combine}
	\begin{adjustbox}{width=0.65\columnwidth, center}
		\begin{tabular}{l|ccc}
			\toprule
			Dataset &  \multicolumn{3}{c}{CIFAR10} \\
			\cmidrule(r){1-1} \cmidrule(l){2-4}
			\# label   &  \multicolumn{1}{c}{40} & \multicolumn{1}{c}{250} & \multicolumn{1}{c}{4000}\\ \cmidrule{1-4}
			FixMatch & 7.47{\scriptsize $\pm$0.28} & 4.86{\scriptsize $\pm$0.05} & 4.21{\scriptsize $\pm$0.08} \\
			FixMatch+FlatMatch & \textbf{6.50}{\scriptsize $\pm$1.25} & \textbf{4.27}{\scriptsize $\pm$2.15} & \textbf{3.92}{\scriptsize $\pm$1.65} \\			
			\midrule
			Dash & 8.93{\scriptsize $\pm$3.11} & 5.16{\scriptsize $\pm$0.23} & 4.36{\scriptsize $\pm$0.11} \\
			Dash+FlatMatch & \textbf{6.73}{\scriptsize $\pm$2.49} & \textbf{4.48}{\scriptsize $\pm$1.56} & \textbf{4.02}{\scriptsize $\pm$1.30} \\			
			\midrule
			FlexMatch & 4.97{\scriptsize $\pm$0.06} & 4.98{\scriptsize $\pm$0.09} & 4.19{\scriptsize $\pm$0.01} \\
			FlexMatch+FlatMatch & \textbf{4.47}{\scriptsize $\pm$0.92} & \textbf{4.25}{\scriptsize $\pm$1.37} & \textbf{3.88}{\scriptsize $\pm$0.75} \\			
			\midrule
			SoftMatch & 4.91{\scriptsize $\pm$0.12} & 4.82{\scriptsize $\pm$0.09} & 4.04{\scriptsize $\pm$0.02} \\
			SoftMatch+FlatMatch & \textbf{4.89}{\scriptsize $\pm$1.32} & \textbf{3.98}{\scriptsize $\pm$1.14} & \textbf{3.84}{\scriptsize $\pm$0.86} \\
			\midrule
			FreeMatch & 4.90{\scriptsize $\pm$0.04} & 4.88{\scriptsize $\pm$0.18} & 4.10{\scriptsize $\pm$0.02} \\
			FlatMatch (Fix label) & \textbf{4.89}{\scriptsize $\pm$1.24} & \textbf{3.90}{\scriptsize $\pm$1.72} & \textbf{3.61}{\scriptsize $\pm$0.49} \\
			\bottomrule
		\end{tabular}
	\end{adjustbox}
\end{table}

\section{Additional Quantitative Results}\label{sec:quantitative}
In this section, we conduct additional experiments on CIFAR10 and ImageNet30~\cite{russakovsky2015imagenet} datasets to compare the performance between some of the most edge-cutting methods, including FixMatch~\cite{sohn2020fixmatch}, Dash~\cite{xu2021dash}, FlexMatch~\cite{zhang2021flexmatch}, FreeMatch~\cite{wang2023freematch}, SoftMatch~\cite{chen2023softmatch}, and our FlatMatch.

\subsection{Combining FlatMatch with Other Methods on CIFAR10}
We choose CIFAR10 dataset with the number of labeled data varied as 40, 250, and 4000, and apply the FlatMatch methodology to several recently proposed SSL methods to show the effectiveness of the proposed cross-sharpness regularization. The results are shown in Table~\ref{tab:combine}, as we can see that our method can further boost the learning performance of all five methods on all three settings, which proves that the cross-sharpness method is quite universal to SSL approaches and can bring non-trivial performance enhancement. Note that in the 40 labels setting, we compute our cross-sharpness on 500 examples with fixed labels, as demonstrated in Section 5.2 from the main paper.

\subsection{Comparing FlatMatch to Other Methods on Large-Scale Datasets and Sophisticated Architecture}
To further testify the performance of FlatMatch on a large-scale datasets, we first conduct experiments on ImageNet30 dataset which is a subset from the original ImageNet dataset and contains 30000 training examples with resolution 256$\times$256 from 30 classes. Moreover, we also test the performance on the original ImageNet dataset. As Vision Transformer (ViT)~\cite{dosovitskiy2020image} has manifested great power on classification tasks, we also adopt ViT as our backbone to validate the performance of FlatMatch.

\begin{table}[!htbp]
	\centering
	\caption{Comparison on ImageNet30.}
	\label{tab:compare}
	\begin{adjustbox}{width=0.45\columnwidth, center}
		\begin{tabular}{l|cc}
			\toprule
			Dataset &  \multicolumn{2}{c}{ImageNet30} \\
			\cmidrule(r){1-1} \cmidrule(l){2-3}
			\# label  & 1500  &  3000 \\ \cmidrule{1-3}
			FixMatch & 12.48{\scriptsize $\pm$0.67} & 8.25{\scriptsize $\pm$0.54}\\
			Dash & 13.29{\scriptsize $\pm$1.26} & 8.79{\scriptsize $\pm$0.42}\\
			FlexMatch & 11.48{\scriptsize $\pm$0.52} & 8.04{\scriptsize $\pm$0.75}\\
			SoftMatch & 10.81{\scriptsize $\pm$0.40} & 7.78{\scriptsize $\pm$0.61}\\
			FreeMatch & 10.34{\scriptsize $\pm$0.46} & 7.21{\scriptsize $\pm$0.19}\\
			FlatMatch & \textbf{9.71}{\scriptsize $\pm$1.55} & \textbf{6.77}{\scriptsize $\pm$1.27}\\
			\bottomrule
		\end{tabular}
	\end{adjustbox}
\end{table}

\begin{table}[!htbp]
	\centering
	\caption{Comparison on ImageNet using Wide-ResNet-28-2 and Vision Transformer.}
	\label{tab:imagenet}
	\begin{adjustbox}{width=\columnwidth, center}
		\begin{tabular}{c|l|lllll}
			\toprule
			\multicolumn{1}{l|}{Dataset} & Architecture     & FlatMatch & FlatMatch-e & FreeMatch & FlexMatch & FixMatch \\ \midrule
			\multirow{2}{*}{ImageNet}    & Wide-ResNet-28-2 & 38.70     & 39.92       & 40.57     & 41.95     & 43.66    \\
			& ViT-Base (86M)   & 21.57     & 22.07       & 23.55     & 23.78     & 25.52    \\ 
			\bottomrule
		\end{tabular}
	\end{adjustbox}
\end{table}

The experiments on ImageNet30 are more time-consuming which normally takes 5 days to finish, much more than CIFAR10 dataset which takes 2 days. We vary the number of labeled data as 1500 and 3000 and show the comparison in Table~\ref{tab:compare}. We observe the effectiveness of FlatMatch over all other baseline methods in both two settings, which again validates the superiority of our method and its effective performance on large-scale datasets.

Additionally, we have conducted the experiments on the original ImageNet dataset by choosing only 100 labels per each class, and provide the test error results of FlatMatch, FlatMatch-e, FixMatch, FlexMatch, and FreeMatch as shown in Table~\ref{tab:imagenet}. We can see that on large-scale dataset such as ImageNet, the performances of both FlatMatch and FlatMatch-e are still superior to other components. Moreover, we can still observe the effectiveness of the two of our methods on ViT. Therefore, it is reasonable to conclude that the performance of FlatMatch is extendable to large-scale datasets and sophisticated architectures.

\begin{figure*}[h]
	\begin{minipage}[t]{0.5\textwidth}
		\centering
		\includegraphics[width=\linewidth]{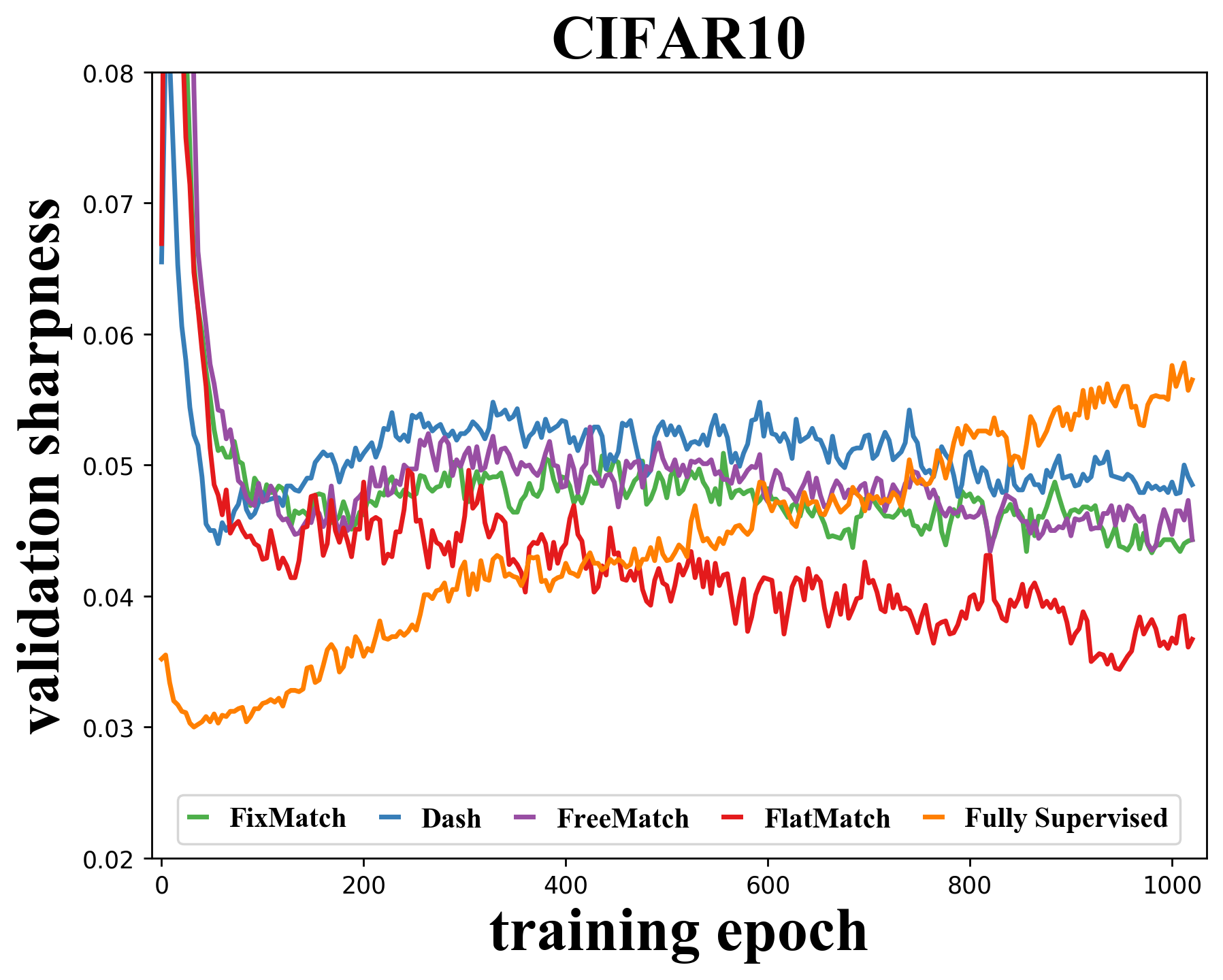}
	\end{minipage}
	\begin{minipage}[t]{0.5\textwidth}
		\centering
		\includegraphics[width=\linewidth]{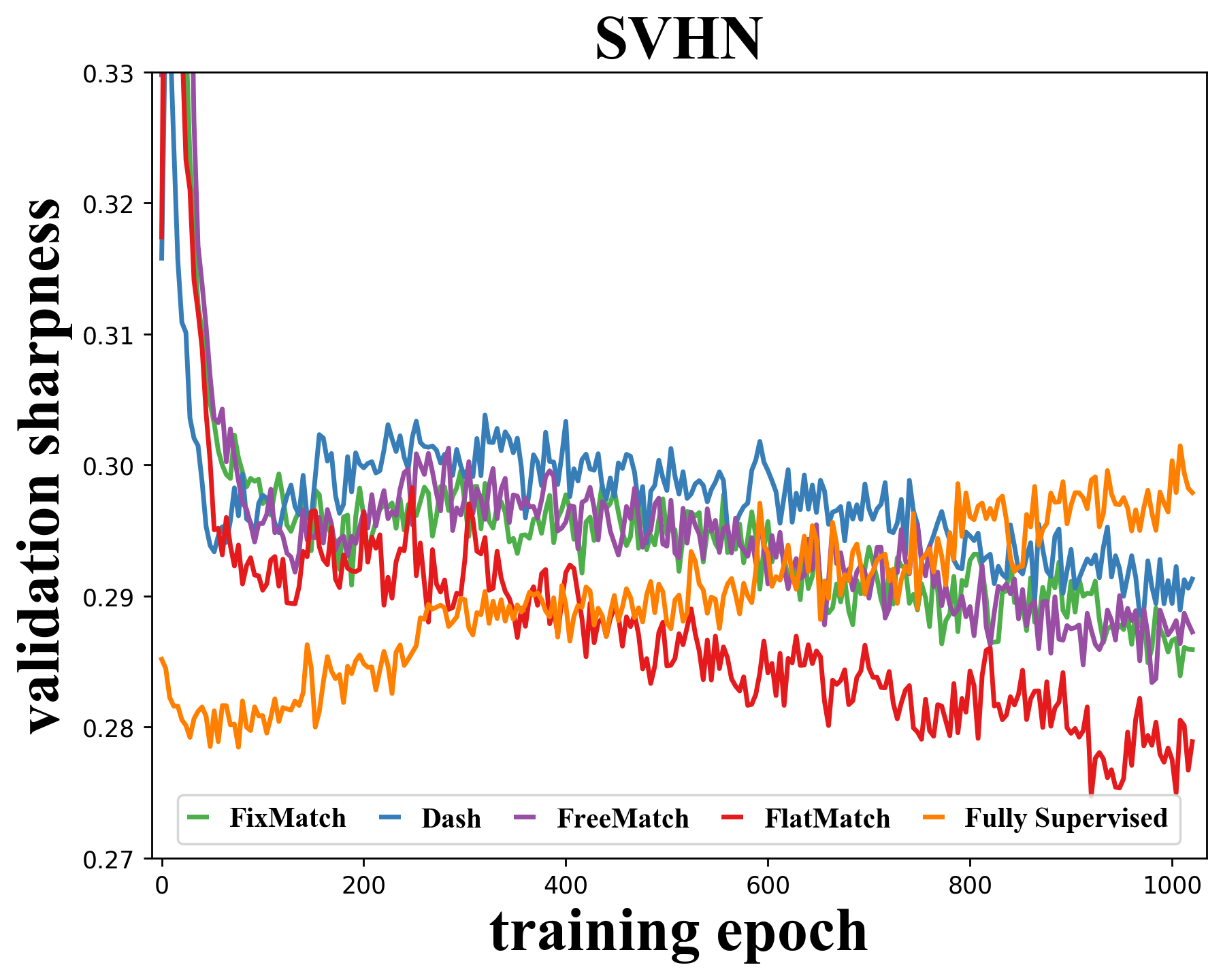}
	\end{minipage}
	\caption{Comparison of sharpness between various SSL methods during training.}
	\label{fig:sharp}
\end{figure*}

\section{Additional Qualitative Results}\label{sec:qualitative}
To further evaluate the flatness of different SSL models during training, we leverage a validation set to compute the sharpness. The sharpness is measured by the increase of loss within a $\ell_2$ bounded neighbor, which is formally defined as $Sharpness:=\mathcal{L}(\theta+\epsilon^*(\theta)) - \mathcal{L}(\theta), \text{where } \epsilon^*(\theta)=\argmax_{\|\epsilon\|_2\le\rho} \mathcal{L}(\theta+\epsilon)$. Specifically, we compare the proposed FlatMatch with FixMatch, Dash, and FreeMatch, and use fully supervised learning as a baseline method. The experiments are conducted on CIFAR10 and SVHN datasets whose results are shown in Figure~\ref{fig:sharp}. First, we observe that FlatMatch achieves the lowest sharpness curve during training on both two datasets, which indicates the SSL model learned by FlatMatch is more robust to perturbations and would not oscillate significantly when facing changes in parameter space. Moreover, we find that fully supervised learning does not improve the flatness as the training proceeds, while all SSL methods can decrease the sharpness to some extent, which demonstrates that training with unlabeled data can help improve the flatness of SSL models.

\begin{figure*}[h]
	\begin{minipage}[t]{0.5\textwidth}
		\centering
		\includegraphics[width=\linewidth]{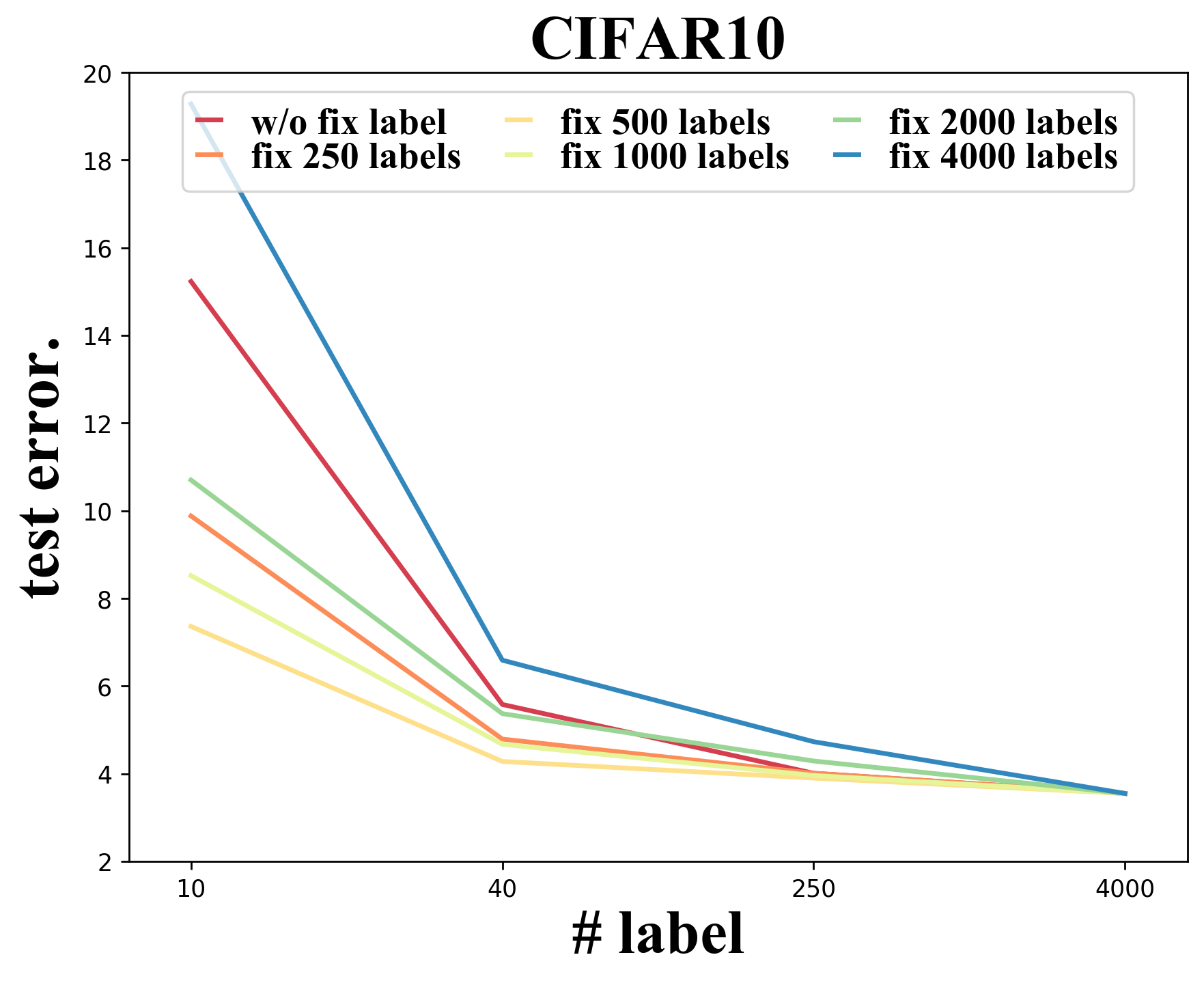}
	\end{minipage}
	\begin{minipage}[t]{0.5\textwidth}
		\centering
		\includegraphics[width=\linewidth]{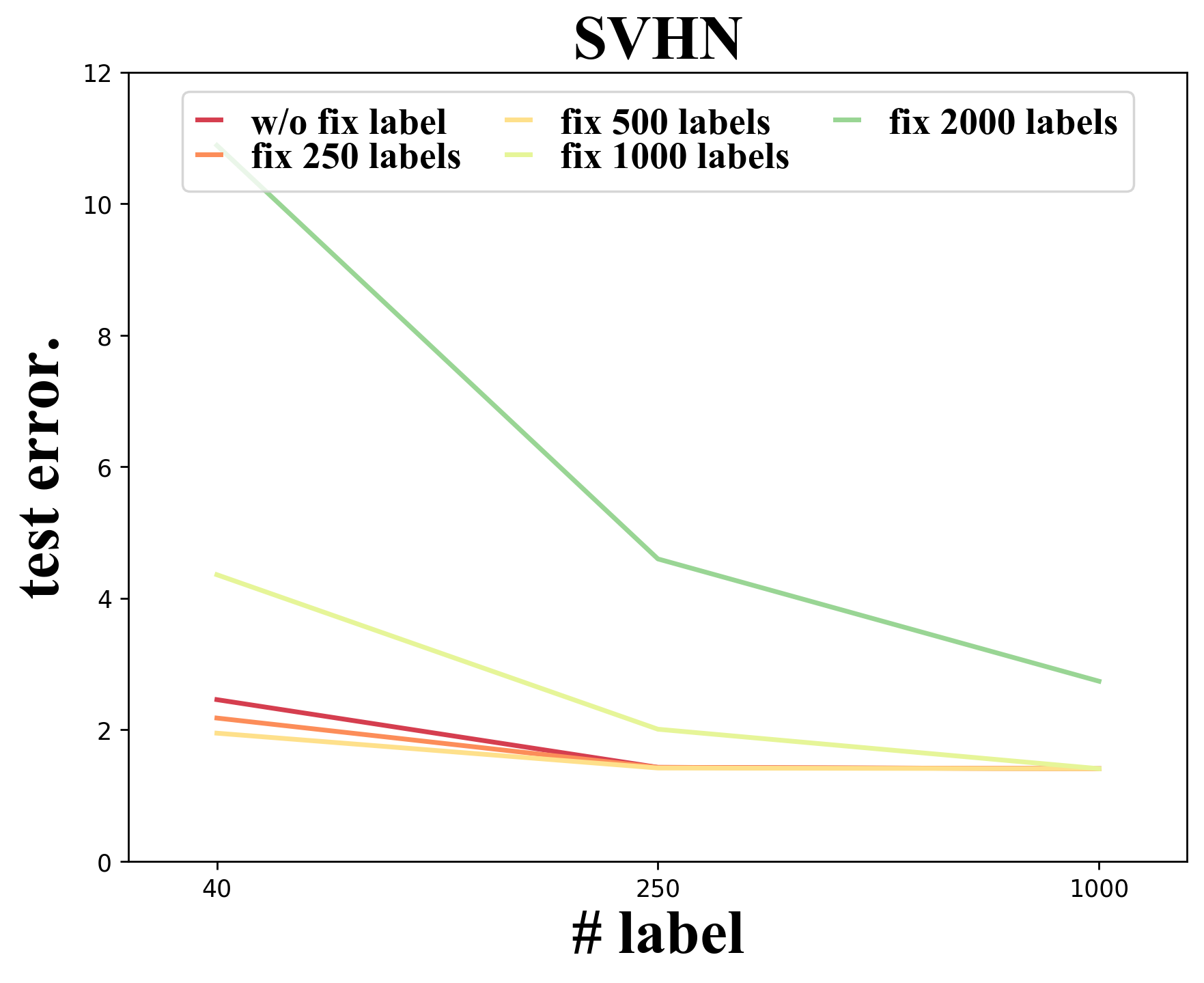}
	\end{minipage}
	\caption{Analysis on changing the number of fixed labels.}
	\label{fig:fixlabel}
\end{figure*}

Furthermore, as shown in the main paper, we find that FlatMatch has limited performance on extremely scare labeled settings. However, this limitation can be addressed by introducing some unlabeled data with fixed labels to improve the computation of cross-sharpness. Hence, here we investigate the effect of changing the number of fixed on the performance of FlatMatch. Specifically, we conduct experiments on CIFAR10 and SVHN datasets and fixing different numbers of labels as 0 (``w/o fix label''), 250, 500, 1000, 2000, 4000\footnote{The 4000 fixed labels setting is not conducted on SVHN as the performance of 2000 fixed labels setting already shows significantly performance degradation.}. The results are shown in Figure~\ref{fig:fixlabel}. We find that both too few fixed labels, \textit{i.e.}, 250 labels and too many fixed labels, \textit{i.e.}, 4000 labels in CIFAR10 and 2000 labels in SVHN, would show a performance drop compared to the optimal number, 500 fixed labels. This is because if the number of fixed labels is too small, the gradient computation would be inaccurate, further limiting the learning results. On the other hand, too many fixed labels would introduce noisy labeled unlabeled data, which would largely mislead the SSL and show serious performance degradation.

\begin{figure*}[h]
	\centering
	\includegraphics[width=0.5\linewidth]{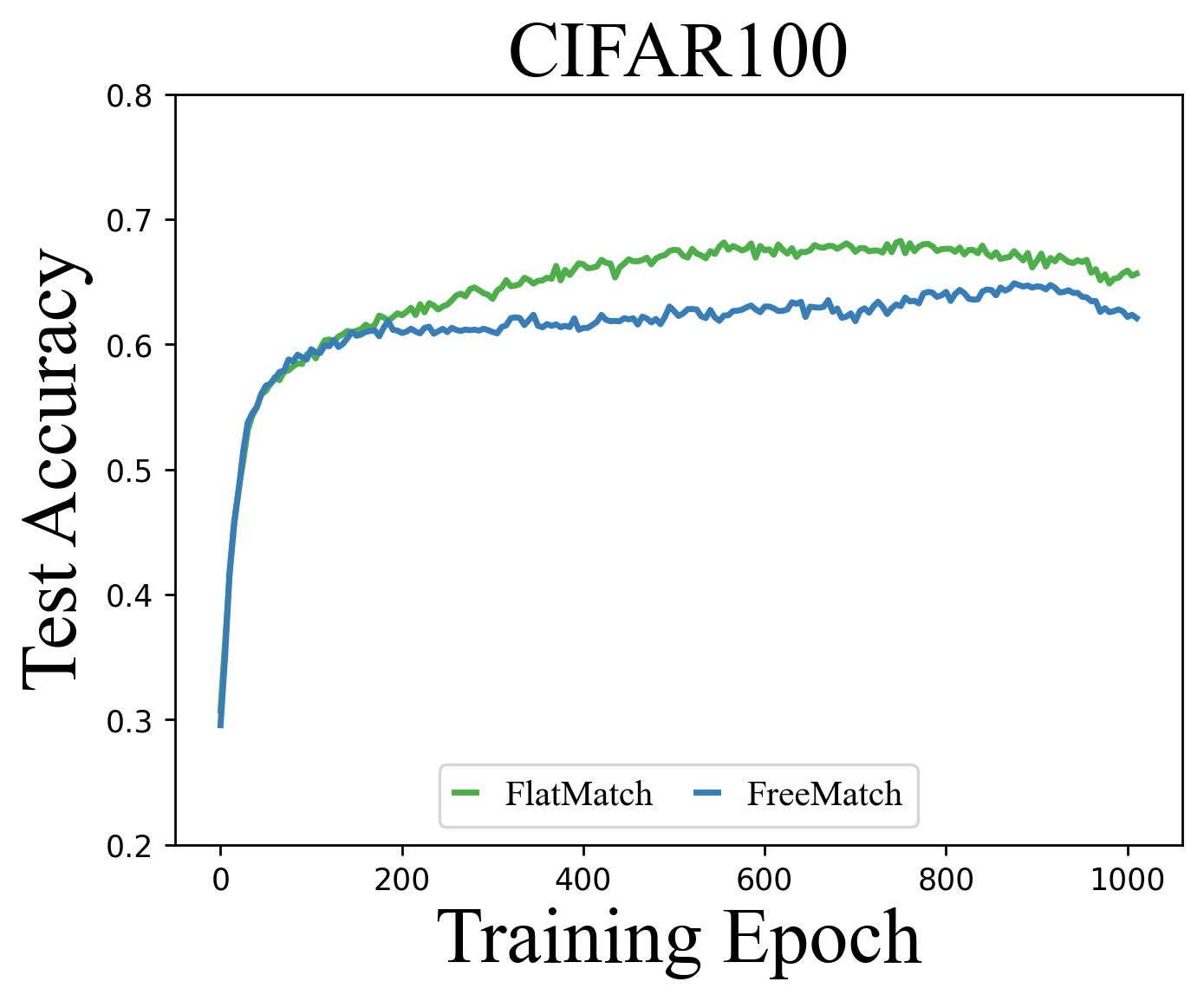}
	\caption{Convergence of accuracy analysis between FlatMatch and FreeMatch.}
	\label{fig:acc_curve}
\end{figure*}

\section{Convergence Analysis}
\label{sec:convergence}
In this section, we conduct analyses regarding the test accuracy and training loss to validate the convergence property of FlatMatch.

For convergence of accuracy, here we compare our FlatMatch with FreeMatch, which has the best performance among most SSL baseline methods. The accuracy curve is shown in Figure~\ref{fig:acc_curve}. We observe that the performances of two methods are almost comparable in early stages, but FlatMatch continues to improve the training performance in the middle and later stages and finally achieves better accuracy than FreeMatch on the final point. Therefore, we can conclude that our method can converge to a superior performance than FreeMatch.

\begin{figure*}[h]
	\begin{minipage}[t]{0.33\textwidth}
		\centering
		\includegraphics[width=\linewidth]{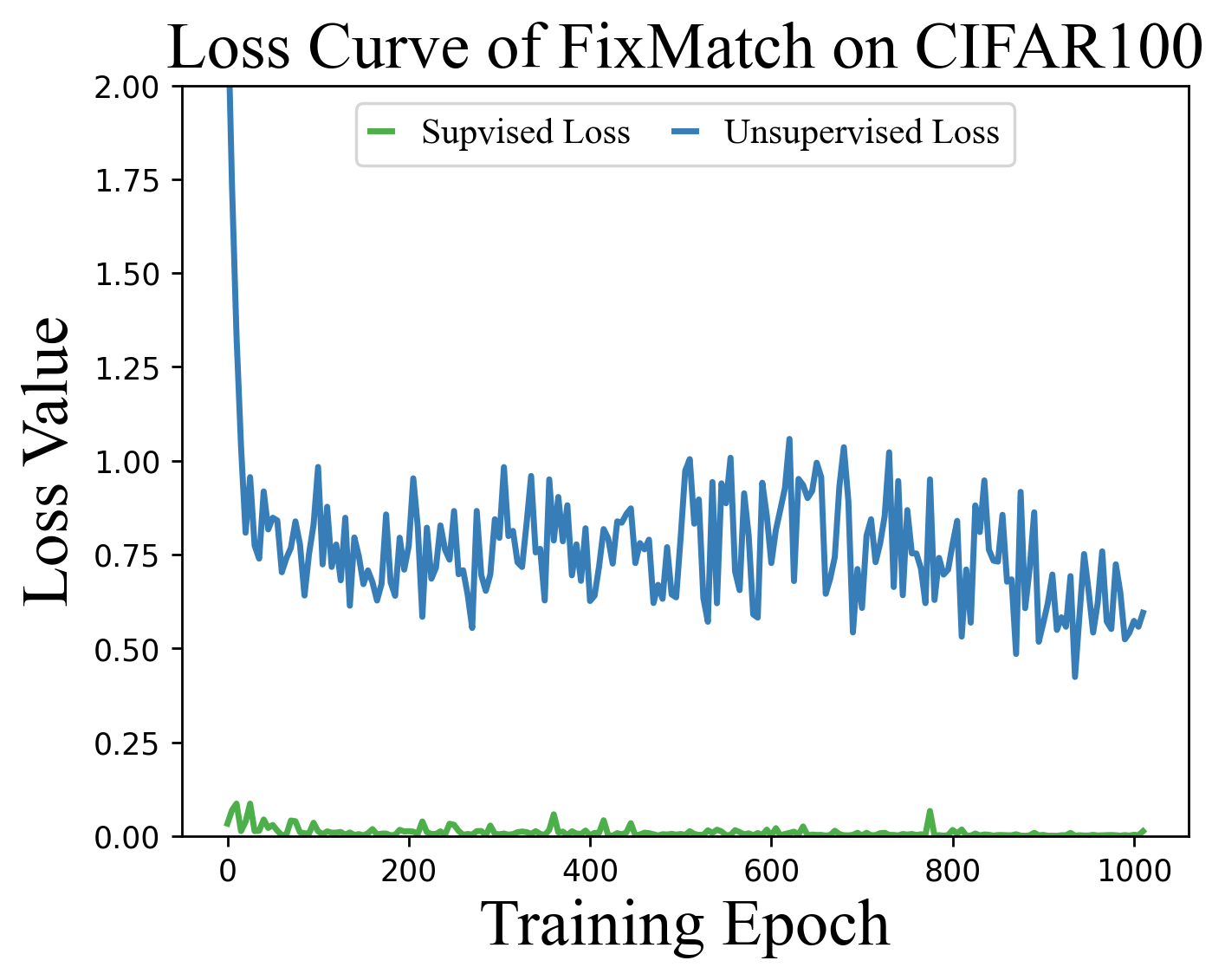}
	\end{minipage}
	\begin{minipage}[t]{0.33\textwidth}
		\centering
		\includegraphics[width=\linewidth]{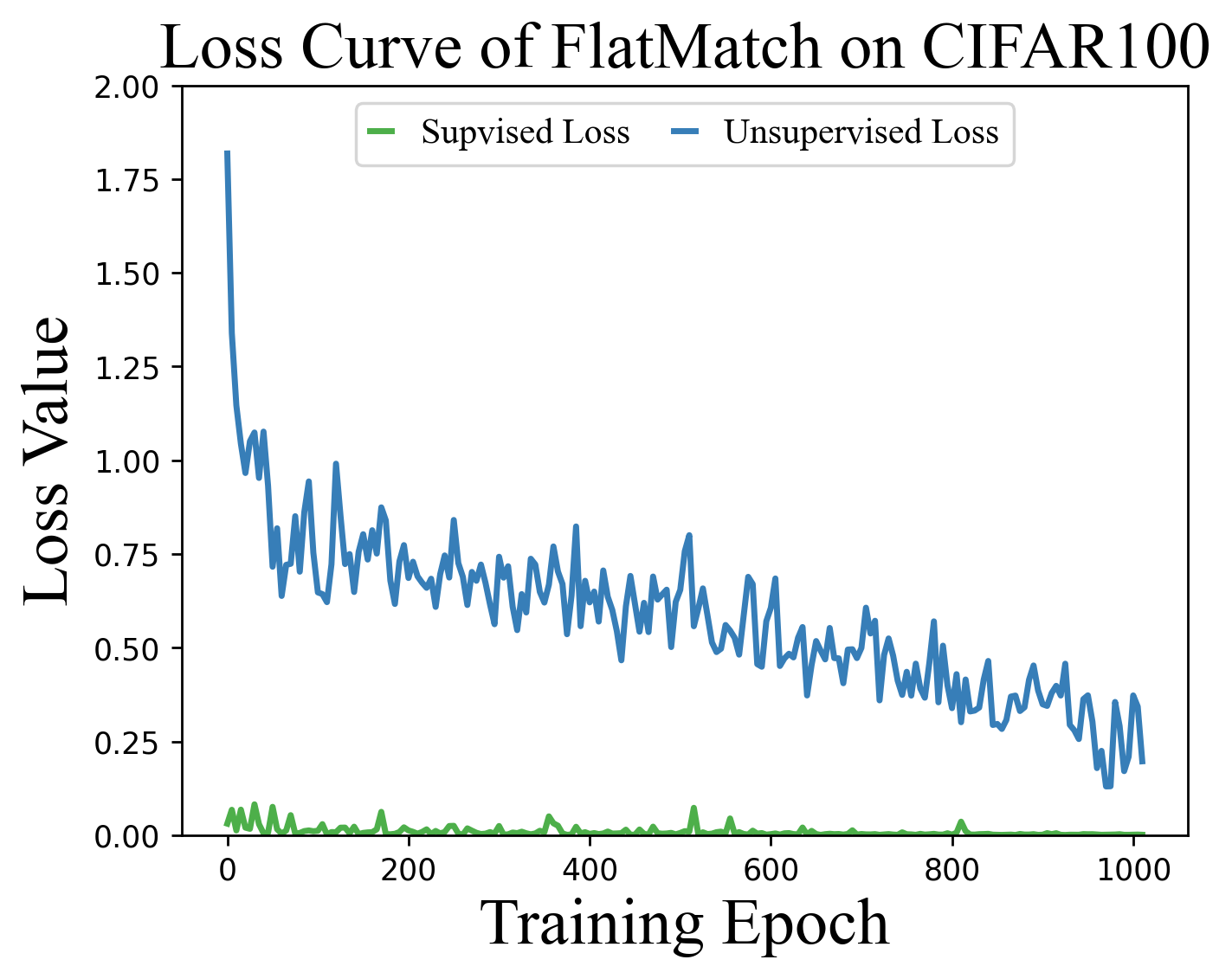}
	\end{minipage}
	\begin{minipage}[t]{0.33\textwidth}
		\centering
		\includegraphics[width=\linewidth]{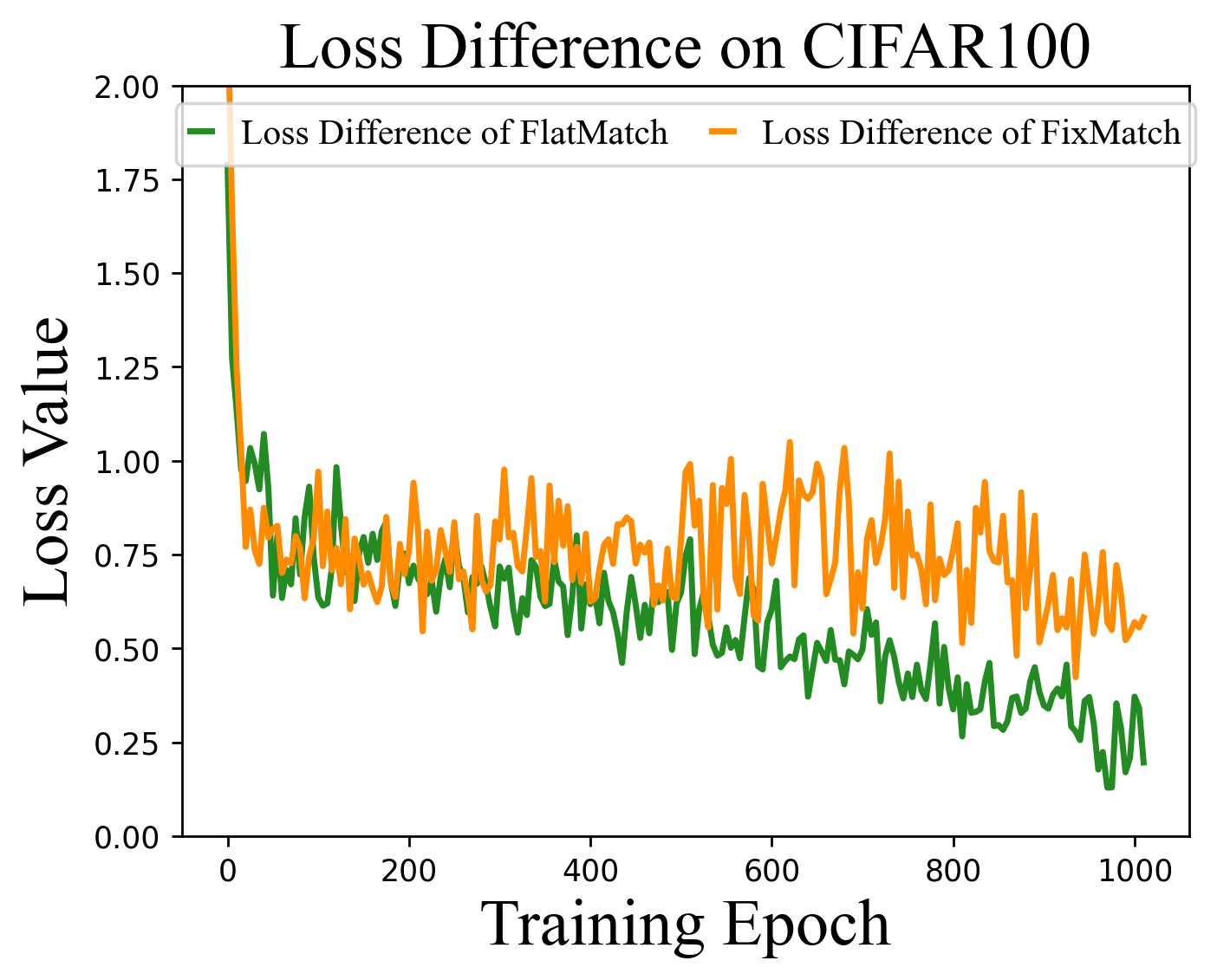}
	\end{minipage}
	\caption{\small Convergence of loss values of FlatMatch and FixMatch.}
	\label{fig:loss_curve}
\end{figure*}

Moreover, to illustrate the convergence of loss curves, we show the loss values of labeled data and unlabeled data from both FlatMatch and FixMatch during training. Moreover, we subtract the loss value of unlabeled data with loss of labeled data to compute a loss difference, which gives an illustration about the performance gap between both two datasets.  

\begin{figure*}[h]
	\begin{minipage}[t]{0.32\textwidth}
		\centering
		\text{\small \ 2D contours of labeled data}
		\includegraphics[width=\linewidth]{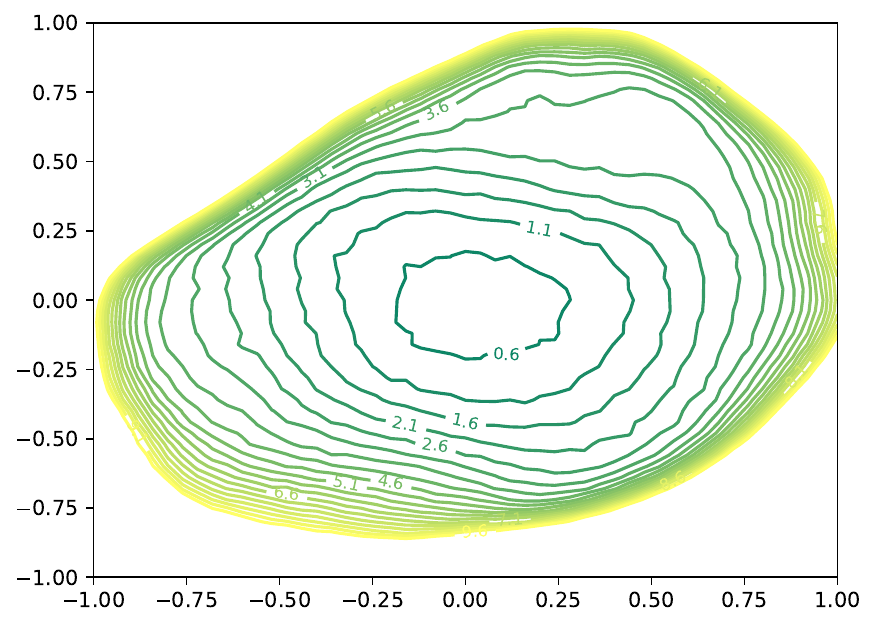}
	\end{minipage}
	\begin{minipage}[t]{0.32\textwidth}
		\centering
		\text{\small \ 2D contours of unlabeled data}
		\includegraphics[width=\linewidth]{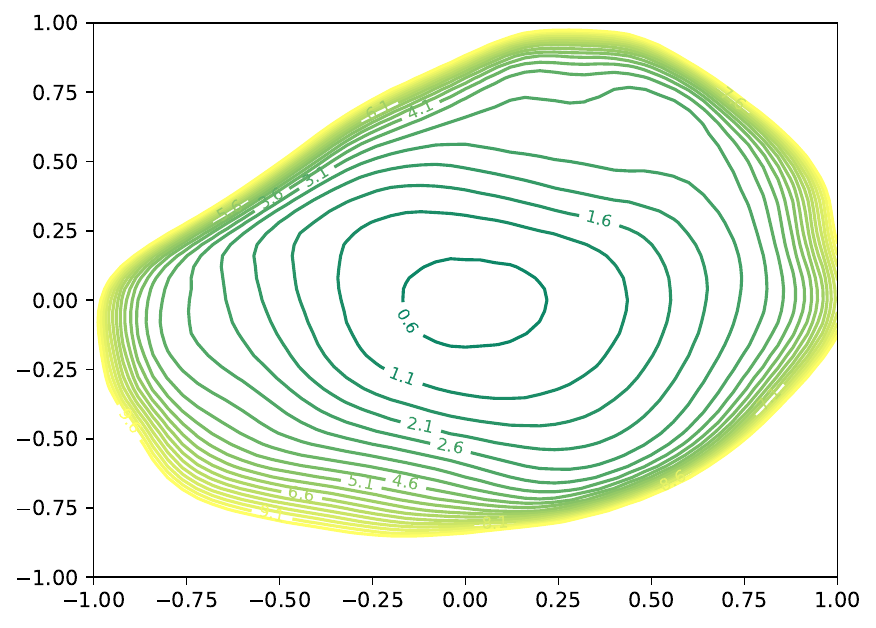}
	\end{minipage}
	\begin{minipage}[t]{0.35\textwidth}
		\centering
		\text{\small1D loss curves}
		\includegraphics[width=\linewidth]{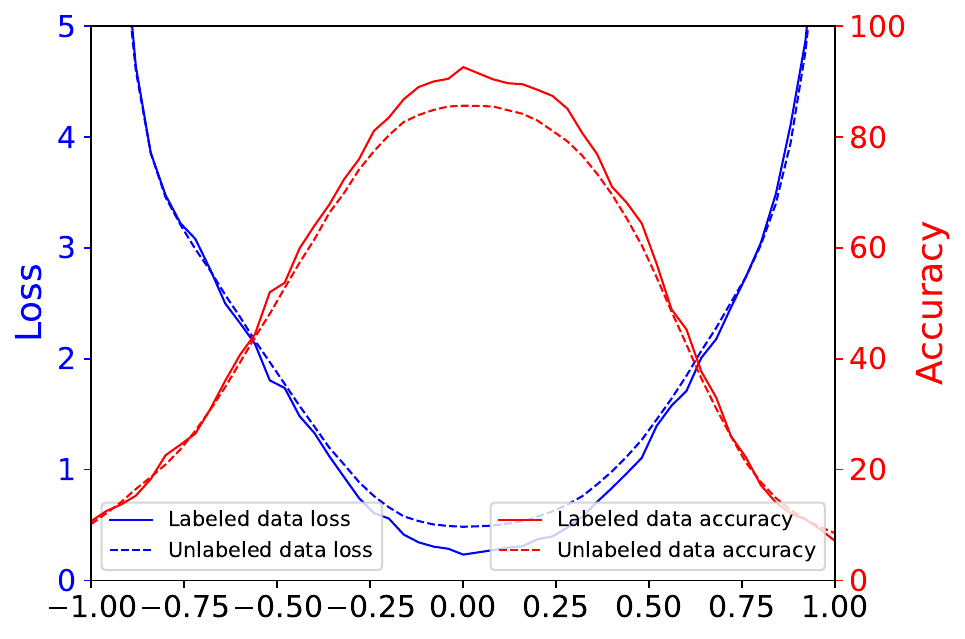}
	\end{minipage}\\
	\text{ \small \qquad\qquad\qquad\qquad\qquad\qquad\qquad\qquad Loss curves of FixMatch}\\
	\vspace{1mm}\\
	\begin{minipage}[t]{0.32\textwidth}
		\centering
		\includegraphics[width=\linewidth]{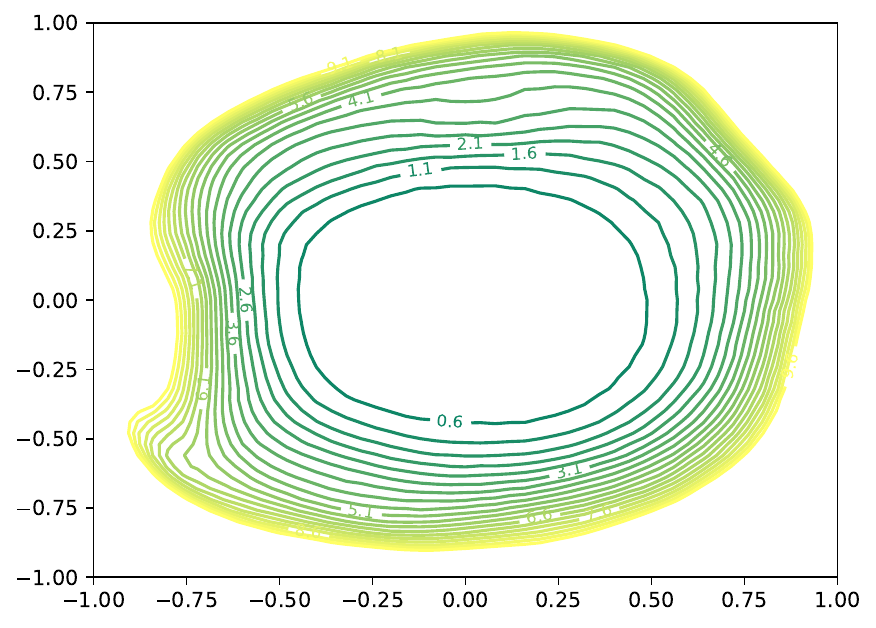}
	\end{minipage}
	\begin{minipage}[t]{0.32\textwidth}
		\centering
		\includegraphics[width=\linewidth]{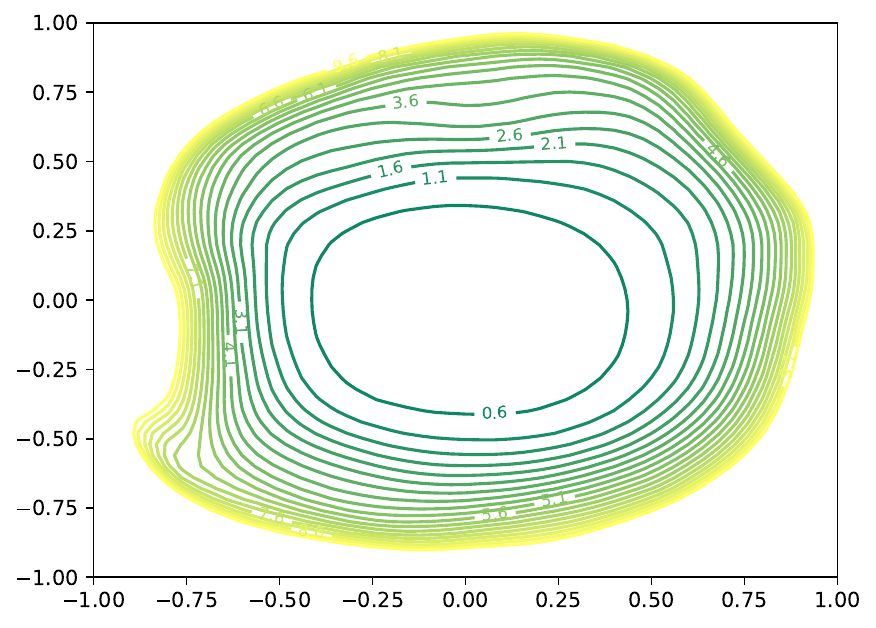}
	\end{minipage}
	\begin{minipage}[t]{0.35\textwidth}
		\centering
		\raisebox{0.01\height}{\includegraphics[width=\linewidth]{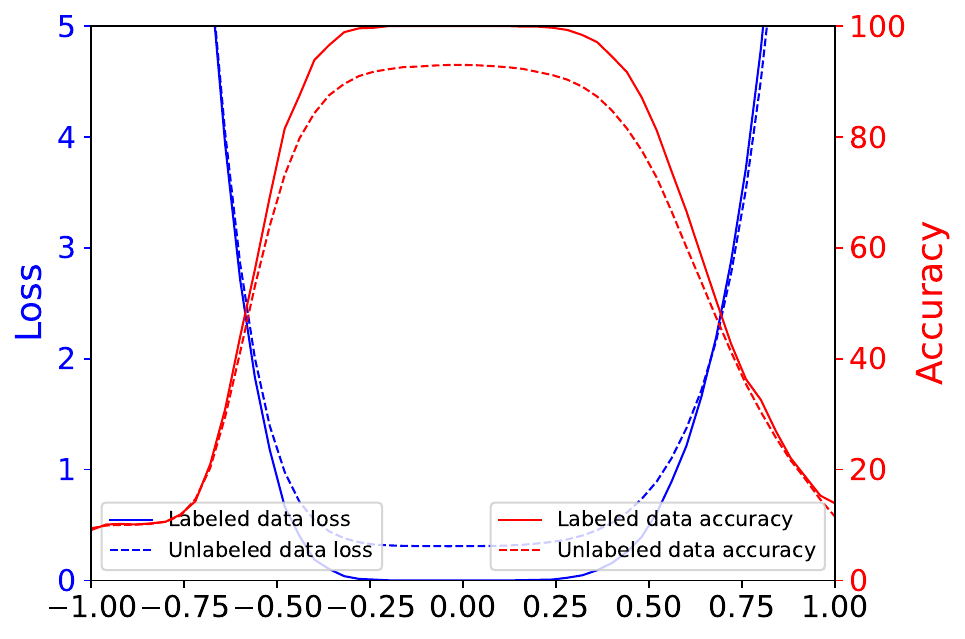}}
	\end{minipage}\\
	\text{ \small \qquad\qquad\qquad\qquad\qquad\qquad\qquad\qquad Loss curves of FlatMatch}\\
	\caption{\small Loss landscapes of labeled data and unlabeled data obtained simultaneously from training using FlatMatch and FixMatch on CIFAR10 with 250 labels per class. The results are generated from the last model checkpoint. The first column and second column show the 2D loss contours of labeled data and unlabeled data, respectively, and the last column shows the 1D loss curves.}
	\label{fig:sup_vis}
\end{figure*}

There are two findings: 1) The loss value of labeled data quickly converges to zero and is significantly smaller than unlabeled data. Such phenomenon occurs in two methods which supports our claim that the learning on labeled data converges faster than unlabeled data. 2) The loss difference between two datasets of FlatMatch is significantly smaller than FixMatch, which indicates that our FlatMatch can alleviate the unmatched convergence speed of two datasets and helps decrease the loss gap between two datasets.

\section{More Visualizations}
\label{sec:more_vis}
To show how the loss curves appears in the later stage of training, we generate the loss landscape of FixMatch and FlatMatch on the last training epoch ($2^{20}$). We can see that FlatMatch generates a wider loss landscape than FixMatch. Moreover, the loss curve of labeled data from FlatMatch is smoother than that of FixMatch. Therefore, we can again conclude that FlatMatch can benefit the generalization result.

\section{Summary and Future Work}\label{sec:summary}
In this paper, we propose a novel FlatMatch approach that minimizes the cross-sharpness measure to improve the generalization performance of SSL. Through extensive quantitative and qualitative experiments, we have thoroughly evaluated the performance of FlatMatch and demonstrated its superiority to other compared methods. Thanks to the generalization improvement of FlatMatch, the classification accuracy on many scenarios have even passed the fully-supervised baseline. 

However, the learning performance of SSL still largely depends on the careful selection of labeled data. Specifically, in the barely-supervised learning scenario, if the selected scarce labeled data deviate from the cluster center, the learning performance of many existing SSL methods would be significantly affected. This is due to the generalization performance between labeled data and unlabeled data being largely mismatched. Under this scenario, the performance of FlatMatch should be further evaluated.


\end{document}